\documentclass{article}
\usepackage{iclr2027_conference,times}

\usepackage{amsmath,amsfonts,bm}

\def\eqref#1{equation~\ref{#1}}

\def\1{\bm{1}}

\DeclareMathAlphabet{\mathsfit}{\encodingdefault}{\sfdefault}{m}{sl}
\SetMathAlphabet{\mathsfit}{bold}{\encodingdefault}{\sfdefault}{bx}{n}

\usepackage[pagebackref=true,breaklinks=true,colorlinks=true,citecolor=NavyBlue,urlcolor=NavyBlue,bookmarks=false]{hyperref}
\usepackage{graphicx}
\usepackage{multirow}
\usepackage[table,dvipsnames]{xcolor}
\usepackage{booktabs}
\usepackage{adjustbox}
\usepackage{capt-of}

\title{EagleDepth: Efficient Fine-Grained Depth Estimation via Pixel Diffusion Decoder}

\author{\begin{minipage}[t]{\dimexpr\textwidth-2\tabcolsep\relax}
\centering
\normalfont\normalsize
\textbf{Bowen Chai}$^{1*}$, \textbf{Tianbao Zhang}$^{1*}$, \textbf{Shuyu Wu}$^{1}$,
\textbf{Dexin Zuo}$^{1}$, \textbf{Zhaoxin Fan}$^{2}$, \textbf{Danping Zou}$^{1\dagger}$\\[0.4em]
\small
$^{1}$Shanghai Jiao Tong University\quad $^{2}$Beihang University\\[0.25em]
\textit{Project Page:}~\href{https://sjtu-visys-team.github.io/EagleDepth/}{sjtu-visys-team.github.io/EagleDepth/}
\end{minipage}}

\renewenvironment{abstract}{
\par\vspace{6pt}\begingroup\setlength{\parskip}{0pt}
\centerline{\large\sc Abstract}
\vspace{5pt}
\begin{list}{}{\setlength{\leftmargin}{3pc}\setlength{\rightmargin}{3pc}\setlength{\topsep}{0pt}\setlength{\partopsep}{0pt}\setlength{\parsep}{0pt}}
\item[]
}{\end{list}\endgroup\vspace{2pt}}

\iclrfinalcopy
\begin{document}

\maketitle
\fancyhead{}
\begingroup
\renewcommand{\thefootnote}{\fnsymbol{footnote}}
\footnotetext[1]{Equal contribution.\quad
  \textsuperscript{\(\dagger\)}Corresponding author:
  Danping Zou, dpzou@sjtu.edu.cn.}
\endgroup
\providecommand{\methodname}{\textsc{EagleDepth}}

\noindent
\begin{minipage}{\textwidth}
    \centering
    \vspace{-4mm}
    \includegraphics[width=\linewidth]{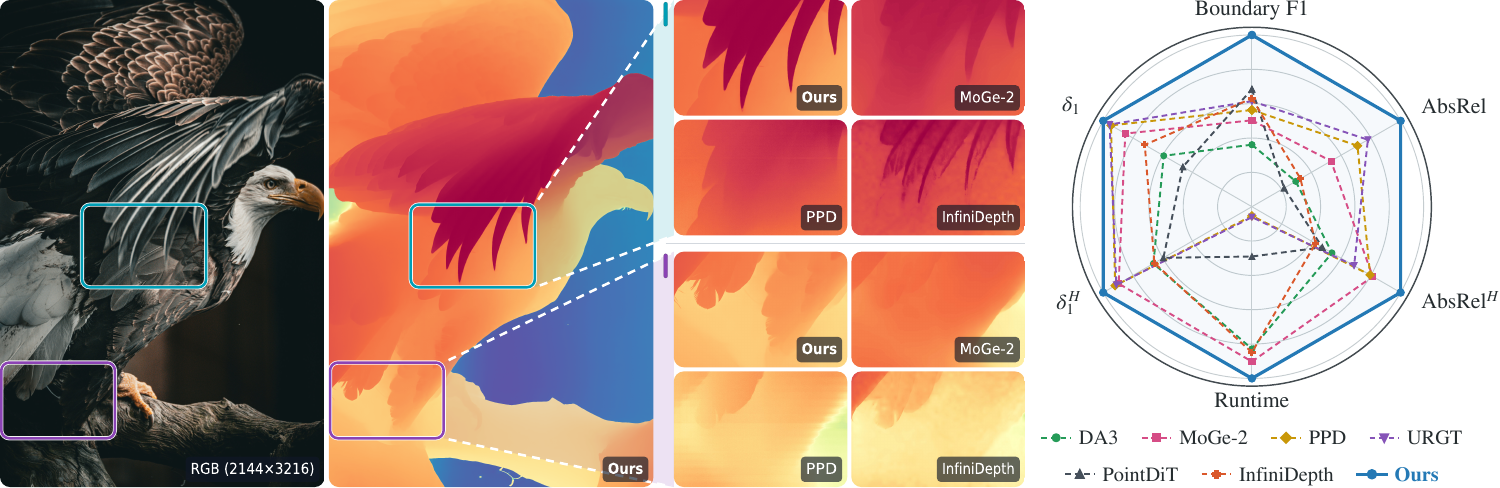}
    \vspace{-6mm}
    \captionof{figure}{\textbf{\methodname{} excels at predicting fine-grained depth maps from high-resolution images.} Compared with state-of-the-art methods, our model achieves better depth estimation results with faster inference, as shown in the radar chart (quality metrics averaged over the five Synth4K~\citep{yu2026infinidepth} subsets; larger radii indicate better performance)
    }
    \label{fig:teaser}
    \vspace{5mm}
\end{minipage}
\par

\begin{abstract}
Recovering detailed geometry from high-resolution images is critical for precise perception of the surroundings and objects. However, existing methods which use latent-space modeling and VAE reconstruction can compromise geometric details. Furthermore, decoding from latent codes introduces substantial inference overhead. To address those issues, we present \methodname{}, an efficient framework for high-resolution monocular depth estimation that combines the geometric priors of latent diffusion with fine-grained pixel-space generation. Our key idea is to retain depth-aware latent representations as guidance while generating the final depth map directly in pixel space. We train the latent and pixel components sequentially: first, we fine-tune a pretrained latent diffusion model using paired RGB--depth supervision; then, we adapt a pretrained pixel diffusion decoder, PiD, to predict depth conditioned on the learned features. Training of the pixel component starts at 1024 resolution and continues across multiple resolutions up to 4K. The latent branch processes resized, lower-resolution RGB images, while the pixel branch generates depth at the target resolution, bypassing the original VAE decoder. This design preserves learned geometric knowledge without requiring the latent backbone to operate at the output resolution. On five commonly used depth estimation datasets and the high-resolution Synth4K dataset, our framework achieves state-of-the-art depth estimation performance, with faster inference and better preservation of fine structures and object boundaries.
\end{abstract}
\vspace{-2mm}
\section{Introduction}
\label{sec:introduction}

Monocular depth estimation provides a fundamental capability for robotic navigation~\citep{vds,flownav}, manipulation~\citep{litemvs,2k,yuan2025depthvla}, and 3D scene understanding~\citep{2,3,4}. Modern cameras capture multi-megapixel images containing small objects, thin structures, and closely spaced boundaries, creating a demand for depth predictions that preserve geometric detail at comparable resolutions. However, representative generative depth and geometry estimators are trained at resolutions such as $512\times512$ and $1024\times768$ pixels~\citep{xu2026pixel,xu2026pointdit}. Downsampling images to these working resolutions can weaken or eliminate visual cues associated with small structures. Efficiently recovering high-resolution, fine-grained geometry is therefore important for applications such as small-obstacle avoidance and precise robotic manipulation.

Preserving local detail and global scene consistency at high resolution requires careful control of computational cost. Interpolation cannot recover unresolved structures, while increasing the working resolution raises computation and memory demands. Methods such as PatchFusion~\citep{patchfusion} and PatchRefiner~\citep{li2026patchrefiner} refine local patches using global image context to recover detailed, coherent depth maps. For diffusion models, iterative denoising can further increase inference costs, particularly when processing large spatial grids.

Generative methods~\citep{hu2025depthcrafter,geowizard} such as Marigold~\citep{marigold} demonstrate that pretrained latent diffusion models can acquire depth-specific knowledge through paired RGB--depth supervision. These representations help resolve ambiguity by providing learned geometric context when visual evidence is incomplete. However, latent compression can obscure narrow structures and closely spaced boundaries, while reconstruction with a variational autoencoder (VAE) can introduce additional geometric errors. Boundary degradation occurs even when reconstructing encoded ground-truth geometry~\citep{xu2026pixel,xu2026pointdit}. Moreover, the decoder must reconstruct the full-resolution output, adding computation as image size increases. These limitations motivate retaining useful latent priors without the original reconstruction pathway.

Pixel-space diffusion avoids VAE reconstruction by generating geometric outputs directly. Pixel-Perfect Depth~\citep{xu2026pixel} predicts depth using pretrained image encoders such as Depth Anything V2, while PointDiT~\citep{xu2026pointdit} generates dense 3D point maps conditioned on frozen DINOv3 image features. However, processing high-resolution spatial representations remains computationally demanding as the number of spatial tokens increases, and iterative sampling can further increase latency. Direct pixel-space generation therefore still faces the challenge of recovering fine geometry efficiently at multi-megapixel resolutions.

We introduce \methodname{}, a framework that combines depth-adapted latent representations with direct pixel-space prediction. Our key idea is to separate geometric prior extraction from high-resolution detail recovery to allocate computation across different resolutions. The latent branch supplies depth-aware guidance at low resolution, while the original RGB image provides the local evidence needed to resolve fine structures. We adapt the pretrained Pixel Diffusion Decoder (PiD)~\citep{lu2026pid}, originally designed to jointly decode and upsample image latents. Depth-specific supervision adapts this formulation to geometric prediction, where details must reflect meaningful surfaces and depth discontinuities rather than visual sharpness alone.

We train the two components sequentially. First, we fine-tune FLUX.2-klein-4B~\citep{bfl2026flux2klein}, a pretrained latent diffusion model, on paired RGB--depth data. This branch processes resized, lower-resolution RGB images to provide depth-aware guidance. We then freeze this branch and fine-tune PiD, conditioned on its guidance and the original RGB image. Pixel training starts with approximately $1024^2$ pixels per image and continues across multiple resolutions up to 4K. The pixel branch directly generates target-resolution depth, bypassing the original VAE decoder during inference without scaling the latent backbone to the output resolution.

We evaluate this design on six benchmark datasets, including the high-resolution Synth4K dataset, covering standard depth estimation and high-resolution prediction up to 4K. \methodname{} achieves state-of-the-art depth estimation performance, with better preservation of fine structures and object boundaries, together with faster inference.

Our contributions can be summarized as follows:
\begin{itemize}
    \item We introduce \methodname{}, a generative monocular depth estimator that brings together the geometric knowledge of depth-adapted latent diffusion and the pixel-space modeling capabilities for efficient, detailed prediction at high resolutions.
    \item We show that depth-aware features from downsampled RGB inputs can guide depth generation at higher resolutions. This enables our framework to retain latent geometric priors without executing the latent backbone at the output resolution or using its VAE decoder.
    \item Experiments on five commonly used depth estimation datasets and the high-resolution Synth4K dataset establish state-of-the-art depth estimation results for our 1024-resolution model and its multi-resolution extension supporting resolutions up to 4K, with consistent improvements in fine-structure recovery and boundary fidelity, alongside faster inference.
\end{itemize}

\section{Related Work}

\subsection{Monocular Depth Estimation}

Discriminative approaches such as DPT~\citep{ranftl2021vision} and Depth Anything V2~\citep{yang2024depth} directly predict depth using pretrained visual representations, while PatchFusion~\citep{patchfusion} extends depth estimation to high resolutions through patch-based fusion. Generative approaches leverage priors learned from image synthesis: Marigold~\citep{marigold} and Lotus-G~\citep{he2025lotus} adapt pretrained latent diffusion models for depth prediction. To address geometric artifacts associated with latent reconstruction, Pixel-Perfect Depth~\citep{xu2026pixel} directly generates depth maps in pixel space, while PointDiT~\citep{xu2026pointdit} applies pixel-space diffusion to dense 3D point maps. These methods demonstrate the value of pretrained representations and direct geometric generation, although efficiently recovering fine-grained geometry at multi-megapixel resolutions remains challenging. Our approach combines these strengths by using depth-adapted features from a lower-resolution latent backbone to guide depth generation at the target resolution.

\subsection{Diffusion Models}
Latent diffusion models~\citep{ldm, magicvideo} reduce generative modeling costs by operating in compressed representations and reconstructing outputs through pretrained autoencoders. Although computationally effective, this formulation makes output fidelity dependent on latent compression and reconstruction. Recent advances revisit diffusion directly in pixel space. JiT~\citep{jit} demonstrates effective clean-image prediction with plain transformers, while PixelDiT~\citep{pixeldit} combines patch-level and pixel-level processing for efficient pixel-space generation. Complementing these approaches, PiD~\citep{lu2026pid} introduces a latent-conditioned pixel diffusion decoder that jointly performs decoding and spatial upsampling, replacing conventional VAE decoding. We build on this pretrained decoder and adapt it from image synthesis to monocular depth estimation. Specifically, depth-aware features learned through latent diffusion adaptation condition pixel-space depth generation, allowing us to reuse pretrained generative knowledge for efficient, high-resolution geometric prediction.

\section{Method}

\subsection{Preliminaries: Flow Matching}

Flow matching learns a time-dependent velocity field that transports a noise distribution to a target data distribution~\citep{lipman2022flow}.
Given a clean sample $\boldsymbol{x}_0$ and Gaussian noise $\boldsymbol{\epsilon}\sim\mathcal{N}(\boldsymbol{0},\boldsymbol{I})$, linear interpolation defines the state and target velocity at time $t\in[0,1]$:
\begin{equation}
    \boldsymbol{x}_t
    =(1-t)\boldsymbol{x}_0+t\boldsymbol{\epsilon},
    \qquad
    \boldsymbol{u}_t
    =\frac{\mathrm{d}\boldsymbol{x}_t}{\mathrm{d}t}
    =\boldsymbol{\epsilon}-\boldsymbol{x}_0,
\end{equation}
with $t=0$ denoting the data endpoint and $t=1$ the noise endpoint. The trainable model can parameterize the velocity field $v_\psi(\boldsymbol{x}_t,t,c)$ directly (v-pred), or express it through a clean-sample predictor $f_\psi$ (x-pred) as
$v_\psi(\boldsymbol{x}_t,t,c)=(\boldsymbol{x}_t-f_\psi(\boldsymbol{x}_t,t,c))/t$, where $c$ denotes conditioning information.
At inference, samples are generated by solving the ODE $\mathrm{d}\boldsymbol{x}_t/\mathrm{d}t=v_\psi(\boldsymbol{x}_t,t,c)$ from $t=1$ to $t=0$.

Pretrained diffusion models provide rich visual priors for depth estimation~\citep{ke2024repurposing}.
Previous studies~\citep{he2025lotus2,he2025lotus,zhang2026dvd} have shown that generative depth estimation models benefit from using RGB images as inputs rather than noise and can produce accurate predictions in a single forward pass.
In line with these findings, we adopt a single-step flow matching formulation that treats the RGB input $\boldsymbol{x}_{\mathrm{rgb}}$ as the state $\boldsymbol{x}_{t^\star}$ at a fixed timestep $t^\star\in(0,1)$.
The depth estimate is obtained through a single update:
\begin{equation}
    \hat{\boldsymbol{x}}_{\mathrm{depth}}
    =\boldsymbol{x}_{\mathrm{rgb}}
    -t^\star v_\psi(\boldsymbol{x}_{\mathrm{rgb}},t^\star,c).
\end{equation}
The same formulation applies in latent space for the latent diffusion model, with $\boldsymbol{x}_{\mathrm{rgb}}$ replaced by the encoded RGB latent $\boldsymbol{z}_{\mathrm{rgb}}$ and the output representing a depth latent.

\begin{figure}[t]
    \centering
    \includegraphics[width=\linewidth]{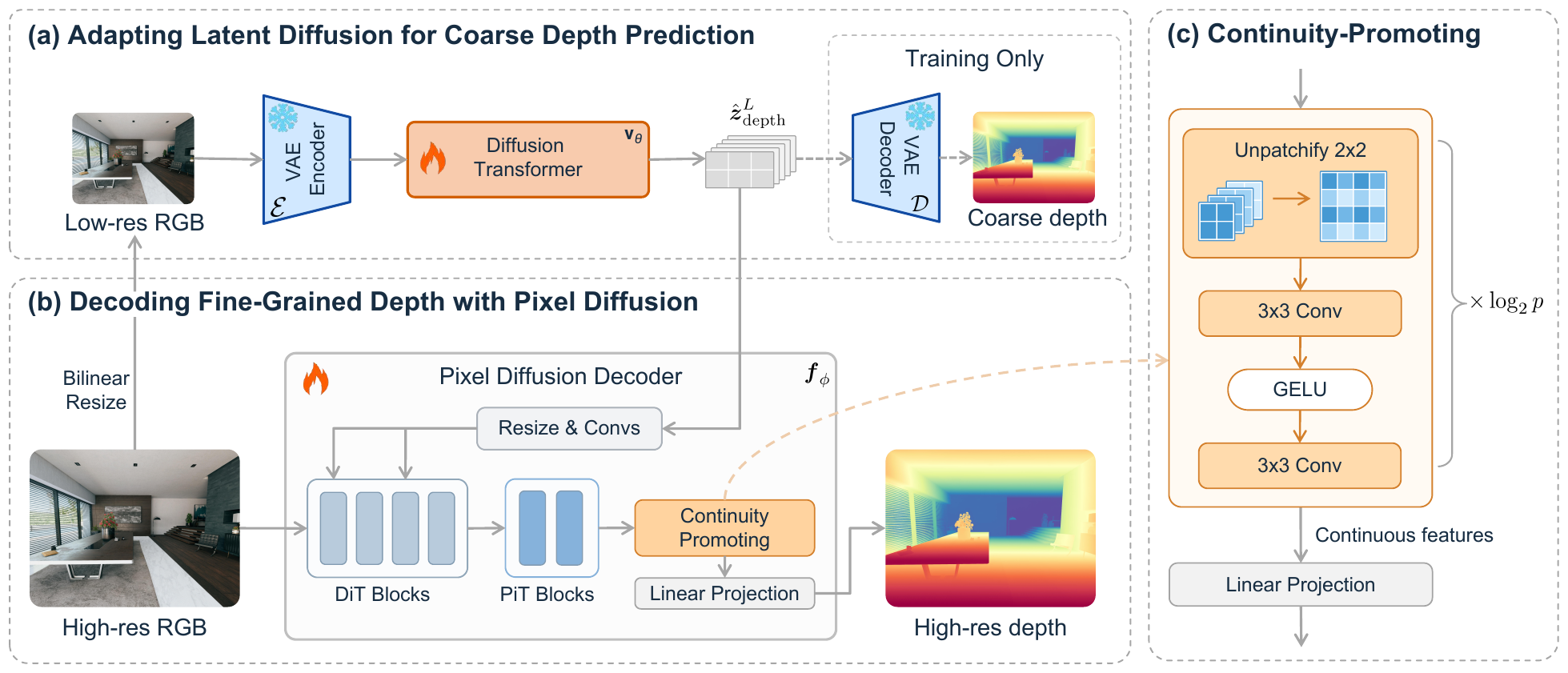}
    \vspace{-6mm}
    \caption{\textbf{Method overview.} \textbf{(a)} An adapted latent predictor, trained with a pixel-space loss, estimates a coarse depth latent from a downsampled RGB image. \textbf{(b)} Conditioned on this latent, the pixel diffusion decoder generates a high-resolution depth map with fine-grained details. \textbf{(c)} A lightweight continuity-promoting module mitigates discontinuity artifacts at patch boundaries.}  
    \label{fig:method}
    \vspace{-2mm}
\end{figure}

\subsection{Framework Overview}
\label{sec:framework_overview}

We aim to produce detailed, high-resolution depth maps, which requires capturing both scene-wide depth relationships and fine local geometry.
Pretrained latent diffusion models combine rich visual priors with compact latent-space modeling.
However, spatial compression and autoencoder reconstruction can lose fine details, and high-resolution VAE decoding incurs substantial computational and memory overhead.
To address these limitations, we draw on the fine-detail modeling capabilities of pixel diffusion.
Specifically, we adapt a pixel diffusion decoder pretrained for image synthesis, PiD~\citep{lu2026pid}, to decode coarse depth latents predicted by a depth-adapted latent diffusion model.
During decoding, the pixel model uses high-resolution RGB observations to recover depth details beyond the coarse representation, producing a detailed depth map at the target resolution.

Figure~\ref{fig:method} illustrates the overall framework of our proposed method.
Given a high-resolution RGB image $\boldsymbol{x}_{\mathrm{rgb}}^{H}\in\mathbb{R}^{H\times W\times3}$, we resize it to a lower resolution (i.e., $512^2$ pixels in total) and encode it using the pretrained autoencoder encoder $\mathcal E$.
The adapted diffusion transformer $v_\theta$ predicts the latent velocity, which is used to obtain a coarse depth latent through a single flow update:
\begin{equation}
    \boldsymbol{z}_{\mathrm{rgb}}^{L}
    =\mathcal E\!\left(\mathcal R_L(\boldsymbol{x}_{\mathrm{rgb}}^{H})\right),
    \qquad
    \hat{\boldsymbol{z}}_{\mathrm{depth}}^{L}
    =\boldsymbol{z}_{\mathrm{rgb}}^{L}
    -t^\star v_\theta(\boldsymbol{z}_{\mathrm{rgb}}^{L},t^\star),
    \label{eq:coarse-depth}
\end{equation}
where $\mathcal R_L$ denotes low-resolution resizing, $t^\star$ is a fixed flow timestep ($t^\star=0.5$ in our method), and $\boldsymbol{z}_{\mathrm{rgb}}^{L},\hat{\boldsymbol{z}}_{\mathrm{depth}}^{L}\in\mathbb{R}^{h_L\times w_L\times C_z}$ denote the RGB and depth latents, respectively.

The pixel diffusion decoder then decodes the predicted coarse depth latent into a high-resolution depth map.
Our pixel decoder backbone, PiD, builds on a pixel diffusion PixelDiT~\citep{pixeldit}, which uses diffusion transformer (DiT) blocks to process patch-level features and pixel transformer (PiT) blocks to refine pixel-level details.
PiD interpolates the latents to be decoded to the target patch grid of size $H/p\times W/p$ and injects them into the DiT blocks through a lightweight convolutional adapter and gated feature addition, where $p$ denotes the patch size.
We retain this conditioning architecture, replacing the image latent with the predicted coarse depth latent and using the original high-resolution RGB image $\boldsymbol{x}_{\mathrm{rgb}}^{H}$ as the pixel diffusion input state.
We further insert a continuity-promoting module before the final linear projection to reduce patch-boundary artifacts in the predicted depth.
The resulting decoder predicts high-resolution depth in a single step:
\begin{equation}
    \hat{\boldsymbol{x}}_{\mathrm{depth}}^{H}
    =
    \Pi\!\left[
        f_\phi\!\left(
            \boldsymbol{x}_{\mathrm{rgb}}^{H},
            t^\star,
            \hat{\boldsymbol{z}}_{\mathrm{depth}}^{L}
        \right)
    \right]
    \in\mathbb{R}^{H\times W},
    \label{eq:pixel-depth}
\end{equation}
where $f_\phi$ is the pixel diffusion decoder and $\Pi$ represents averaging the three output channels into a single depth map.
This coarse-to-fine design enables efficient high-resolution depth prediction with accurate scene-level structure and fine geometric detail.

\subsection{Adapting Latent Diffusion for Coarse Depth Prediction}

Accurate coarse depth provides reliable geometric guidance for subsequent detail refinement, motivating us to improve the accuracy of the latent depth predictor.
Prior works have established effective strategies for adapting pretrained generative models to depth estimation~\citep{ke2024repurposing,he2025lotus2,he2025lotus}.
Motivated by these findings, we adopt deterministic single-step prediction and replace noise inputs with encoded RGB images.
We use FLUX.2-klein-4B~\citep{bfl2026flux2klein}, a pretrained rectified-flow transformer, as our backbone to balance model capability and computational efficiency.
To improve accuracy, we revisit the supervision objective for the latent predictor.

Conventional latent flow matching trains models to predict instantaneous velocities in latent space.
However, flow trajectories are not necessarily straight, so even accurate instantaneous velocities may yield endpoint errors in a single-step update~\citep{frans2025one}.
Moreover, although latent-space supervision encourages predictions to match targets, the decoder may still amplify small residual errors.
Thus, latent-space supervision may be suboptimal.
Since our single-step formulation does not require iterative latent-space sampling, we can directly supervise the decoded depth map:
\begin{equation}
    \hat{\boldsymbol{x}}_{\mathrm{depth}}^{L}
    =
    \Pi\!\left[
        \mathcal D\!\left(\hat{\boldsymbol{z}}_{\mathrm{depth}}^{L}\right)
    \right],
    \qquad
    \mathcal L_{\mathrm{coarse}}
    =
    \mathcal L_{\mathrm{depth}}\!\left(
        \hat{\boldsymbol{x}}_{\mathrm{depth}}^{L},
        \boldsymbol{x}_{\mathrm{depth}}^{L}
    \right),
    \label{eq:coarse-depth-supervision}
\end{equation}
where $\mathcal D$ denotes the VAE decoder, $\Pi$ averages the output channels, $\boldsymbol{x}_{\mathrm{depth}}^{L}$ is the normalized low-resolution ground-truth depth, and $\mathcal{L}_{\mathrm{depth}}$ denotes the loss used for depth supervision.
We freeze the decoder and backpropagate the depth loss through it, training the latent predictor under a fixed decoding mapping.
This objective evaluates the quality of latent predictions by the accuracy of their decoded depth maps, encouraging accurate depth values and geometric structure.

\subsection{Decoding Fine-Grained Depth with Pixel Diffusion}

Central to our approach is adapting a pretrained pixel diffusion decoder to depth estimation.
Compared with a conventional VAE decoder trained for reconstruction, a pixel diffusion decoder provides a generative prior learned in pixel space that can help recover fine details beyond the coarse latent representation.
Specifically, we build upon PiD~\citep{lu2026pid}, which formulates latent decoding as latent-conditioned image generation in pixel space.

While the original model decodes natural-image latents, we fine-tune it to decode the predicted coarse depth latents into detailed, high-resolution depth maps using the high-resolution RGB input, as described in Section~\ref{sec:framework_overview}.
Following prior findings on pixel-space diffusion~\citep{jit}, we use direct clean-sample prediction (x-pred).
To initialize the x-pred decoder from PiD's pretrained velocity model, we negate the weights and bias of its final linear layer: the negated velocity target, $-\boldsymbol{u}_t=\boldsymbol{x}_0-\boldsymbol{\epsilon}$, differs from the clean-sample target $\boldsymbol{x}_0$ only by a noise term.

The unpatchification head in diffusion transformers maps each patch token to a $p\times p$ pixel block without local spatial mixing, leaving discontinuities along patch boundaries.
In depth maps, these artifacts can be especially noticeable at object boundaries, where one patch spans sharply different depths.
Convolutions can mitigate these inconsistencies through local feature fusion~\citep{he2025lotus2,wang2026vggt}.
However, the large patch size in pixel diffusion makes it costly for convolutional heads to capture patch-wide context after unpatchification. We therefore insert a continuity-promoting module before the original output projection, interleaving progressive unpatchification with small-kernel convolutions to mix hidden features at multiple spatial scales.

Specifically, given the normalized hidden features $\boldsymbol{h}^{(0)}$, the module progressively unfolds their within-patch dimensions into the spatial grid and applies convolutions after each unfolding step:
\begin{equation}
    \boldsymbol{h}^{(k+1)}
    =
    \operatorname{Conv}_{3\times3}
    \!\left(
        \operatorname{GELU}\!\left(
            \operatorname{Conv}_{3\times3}
            \!\left(
                \operatorname{Unpatchify}_{2\times2}\!\left(\boldsymbol{h}^{(k)}\right)
            \right)
        \right)
    \right),
    \quad k=0,\ldots,\log_2p-1.
\end{equation}
Each step doubles the spatial grid dimensions until the full pixel resolution is reached.
The original linear projection is then applied to produce the output.
Each Conv--GELU--Conv block is jointly initialized to implement an identity mapping, preserving the pretrained readout at initialization.

\subsection{Training Strategy}
\label{sec:training_strategy}

We train the two components sequentially.
First, we adapt the latent diffusion model for coarse depth prediction at approximately $512^2$ pixels, learning to provide reliable scene-level geometry for subsequent detail refinement.
We then freeze this model and fine-tune PiD at approximately $1024^2$ pixels, adapting its pretrained generative prior to decode the predicted depth latents and recover fine geometric details with RGB guidance. This yields our Ours (1024) variant.

\begin{figure}[t]
    \centering
    \includegraphics[width=\linewidth]{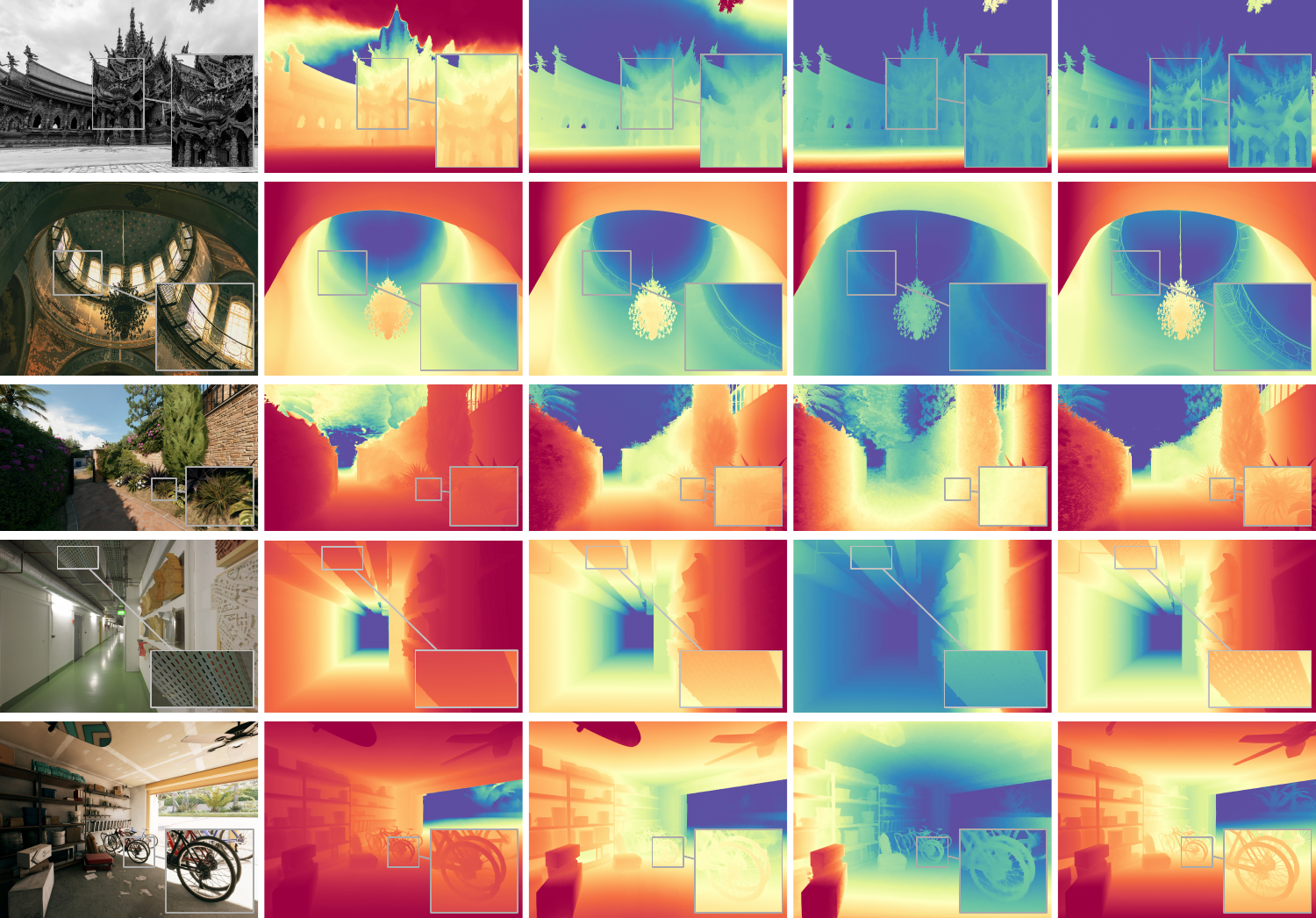}
    \vspace{2pt}
    {\footnotesize
    \makebox[\linewidth]{\makebox[0.196\linewidth]{RGB}\hfill
        \makebox[0.196\linewidth]{MoGe-2}\hfill
        \makebox[0.196\linewidth]{PPD}\hfill
        \makebox[0.196\linewidth]{InfiniDepth}\hfill
        \makebox[0.196\linewidth]{Ours}}}
    \vspace{-6mm}
    \caption{\textbf{Qualitative comparison with leading monocular depth estimation methods.} Our method preserves fine geometric details and thin structures while maintaining sharp depth boundaries and smooth surfaces, as highlighted by the arrows and zoomed-in regions.}
    \vspace{-5mm}
    \label{fig:main_visual}
\end{figure}

To further extend the model's high-resolution capability, we continue training PiD with mixed resolutions up to 4K.
During this training, 25\% of batches follow the same $1024^2$ data settings as before, while the remaining 75\% are drawn exclusively from high-resolution datasets across multiple resolution buckets.
Retaining the $1024^2$ training distribution helps preserve the model's existing capability as it learns to recover details at higher resolutions. This mixed-resolution training yields our Ours (MR) variant.
Throughout PiD training, the latent model remains frozen and processes inputs at approximately $512^2$ pixels regardless of PiD's output resolution, while its predicted latents are resized to the target patch grid before injection into PiD.

Both components are optimized using an L1 loss, a gradient matching loss~\citep{yang2024depth}, and a virtual normal loss~\citep{yin2019enforcing}, with respective weights of $1$, $2$, and $0.5$.

\section{Experiments}

\begin{table}[t]
\centering
\small
\setlength{\tabcolsep}{2.5pt}
\renewcommand{\arraystretch}{1.08}
\setlength{\belowcaptionskip}{6pt}
\vspace{-3mm}
\caption{\textbf{Quantitative results on standard depth benchmarks.} \textbf{Bold} and \underline{underlined} values denote the best and second-best results, respectively. Our method achieves the best results on four of the five benchmarks and remains competitive on KITTI.}
\vspace{-1mm}
\label{tab:comparison-standard}

\begin{tabular*}{\linewidth}{@{\extracolsep{\fill}}l*{10}{c}@{}}
\toprule
\multirow{2}{*}{\textbf{Method}}
& \multicolumn{2}{c}{\textbf{NYUv2}}
& \multicolumn{2}{c}{\textbf{KITTI}}
& \multicolumn{2}{c}{\textbf{ETH3D}}
& \multicolumn{2}{c}{\textbf{ScanNet}}
& \multicolumn{2}{c}{\textbf{DIODE}} \\
\cmidrule(lr){2-3}
\cmidrule(lr){4-5}
\cmidrule(lr){6-7}
\cmidrule(lr){8-9}
\cmidrule(lr){10-11}
& AbsRel $\downarrow$ & $\delta_1\uparrow$
& AbsRel $\downarrow$ & $\delta_1\uparrow$
& AbsRel $\downarrow$ & $\delta_1\uparrow$
& AbsRel $\downarrow$ & $\delta_1\uparrow$
& AbsRel $\downarrow$ & $\delta_1\uparrow$ \\
\midrule
MiDaS & 11.1 & 88.5 & 23.6 & 63.0 & 18.4 & 75.2 & 12.1 & 84.6 & \textemdash & \textemdash \\
DA2 & 4.5 & 97.9 & 7.4 & 94.6 & 13.1 & 86.5 & 6.5 & 97.2 & 6.6 & 95.2 \\
Marigold & 5.5 & 96.4 & 9.9 & 91.6 & 6.5 & 96.0 & 6.4 & 95.1 & 10.0 & 90.7 \\
Lotus-2 & 5.0 & 97.6 & 6.6 & 95.7 & 5.1 & 97.5 & 5.2 & 97.3 & 6.1 & 95.2 \\
DA3 & 4.1 & 98.0 & 5.9 & 97.2 & 4.4 & 97.8 & 3.9 & 97.9 & 5.9 & 95.1 \\
MoGe-2 & \textbf{3.6} & 98.3 & \textbf{5.2} & \textbf{97.9} & 3.5 & 98.7 & \textbf{3.5} & \textbf{98.3} & 4.7 & 96.4 \\
PPD & 3.8 & \underline{98.4} & \textbf{5.2} & \underline{97.4} & 3.2 & \underline{99.1} & 3.8 & 98.2 & 4.1 & 97.7 \\
PointDiT & 6.0 & 96.8 & 11.6 & 88.7 & 6.1 & 96.7 & 5.6 & 97.3 & 6.8 & 94.8 \\
InfiniDepth & 4.8 & 97.4 & 5.8 & 96.0 & 4.5 & 98.0 & 4.9 & 97.0 & 5.2 & 96.4 \\
\midrule
\cellcolor{black!5}\textbf{Ours (1024)}
& \textbf{3.6} & \textbf{98.5} & 5.4 & \underline{97.4}
& \textbf{3.1} & \textbf{99.2} & \textbf{3.5} & \textbf{98.3}
& \textbf{3.9} & \textbf{97.9} \\
\cellcolor{black!5}\textbf{Ours (MR)}
& \textbf{3.6} & \underline{98.4} & 5.4 & 97.3
& \textbf{3.1} & \underline{99.1} & \textbf{3.5} & \textbf{98.3}
& \textbf{3.9} & \textbf{97.9} \\
\bottomrule
\end{tabular*}
\vspace{-4mm}
\end{table}

\subsection{Experimental Setup}

\textbf{Datasets.} Our model is trained on 536K RGB–depth pairs sampled from several datasets: Hypersim~\citep{roberts2021hypersim}, VKITTI2~\citep{cabon2020virtual}, TartanAir~\citep{wang2020tartanair}, IRS~\citep{wang2021irs}, EDEN~\citep{le2021eden}, RobbyReal~\citep{lingbot-depth2026}, UnrealStereo4K~\citep{zhang2018unrealstereo}, UrbanSyn~\citep{gomez2025all}, Spring~\citep{mehl2023spring}, and Rawmantic3D~\citep{rawmantic2026data}. Only the last four datasets are used as high-resolution training dataset. We evaluate depth estimation performance on five real-world benchmarks: NYUv2~\citep{silberman2012indoor}, KITTI~\citep{geiger2012we}, ETH3D~\citep{schops2017multi}, ScanNet~\citep{dai2017scannet}, and DIODE~\citep{vasiljevic2019diode}. To further assess high-resolution prediction and fine-detail recovery, we use Synth4K~\citep{yu2026infinidepth}, a synthetic high-resolution benchmark that comprises five subsets.

\textbf{Implementation details.}
We transform the ground-truth depth into log space via $\log(1+d)$ and normalize each depth map to $-1,1$ using its 2nd and 98th percentiles.
We adapt FLUX.2-klein-4B~\citep{flux-2-2025} with rank-256 LoRA for 30,000 steps at a learning rate of $1\times10^{-4}$. With the latent model frozen, we fully fine-tune PiD~\citep{lu2026pid} at approximately $1024^2$ pixels for 50,000 steps with a learning rate of $3\times10^{-5}$, obtaining Ours (1024).
Mixed-resolution training continues for another 20,000 steps with the learning rate reduced to $10^{-5}$, producing Ours (MR).
All models are trained on 8 NVIDIA GPUs, using the AdamW optimizer and a total batch size of 16.

\textbf{Evaluation protocol.}
We evaluate all comparison methods under the same resolution settings for fair comparisons.
Specifically, images from the five real-world benchmarks are resized to approximately $1024^2$ pixels with aspect ratios preserved, whereas Synth4K is evaluated at native 4K resolution.
We report AbsRel and $\delta_1$ to assess overall depth accuracy.
For datasets with dense ground-truth depth, we further report HF-masked AbsRel and $\delta_1$~\citep{yu2026infinidepth} and boundary F1-score ($r=1$)~\citep{bochkovskiy2025depth} to assess fine-detail recovery in the predicted depth maps.

\subsection{Comparison with State-of-the-Art Methods}

We compare \methodname{} with several leading methods for depth estimation, including: MiDaS~\citep{ranftl2020towards}, DA2~\citep{yang2024depth}, Marigold~\citep{marigold}, Lotus-2~\citep{he2025lotus2}, DA3~\citep{lin2025depth}, MoGe-2~\citep{wang2026moge}, PPD~\citep{xu2026pixel}, PointDiT~\citep{xu2026pointdit}, InfiniDepth~\citep{yu2026infinidepth},  and URGT~\citep{cui2026any}, where the last two methods are designed for high-resolution depth prediction.

\textbf{Qualitative results.}
We present visual comparisons across diverse scenes and image resolutions in Figure~\ref{fig:main_visual}. \methodname{} consistently captures the overall scene structure while preserving local geometric details. Compared with other methods, our model more faithfully recovers fine structures and delineates object boundaries that are otherwise blurred or lost. These visual improvements agree with the gains in HF-masked accuracy and boundary F1, supporting the effectiveness of combining coarse depth guidance with RGB-guided pixel decoding across scenes and resolutions.

\begin{table}[t]
\centering
\small
\setlength{\tabcolsep}{2.5pt}
\renewcommand{\arraystretch}{1.08}
\setlength{\belowcaptionskip}{6pt}
\vspace{-3mm}
\caption{\textbf{Depth accuracy on the high-resolution Synth4K benchmark,} evaluated over the full valid region (Overall) and high-frequency regions. Results for InfiniDepth$^\dagger$ are reported with a $504 \times 896$ input and a 4K output, following its original paper. All other results are reported with 4K inputs and outputs. Our method demonstrates leading performance at high resolution.}
\label{tab:comparison-highres}

\begin{tabular*}{\linewidth}{@{\extracolsep{\fill}}cl*{10}{c}@{}}
\toprule
& \multirow{2}{*}{\textbf{Method}}
& \multicolumn{2}{c}{\textbf{Synth4K-1}}
& \multicolumn{2}{c}{\textbf{Synth4K-2}}
& \multicolumn{2}{c}{\textbf{Synth4K-3}}
& \multicolumn{2}{c}{\textbf{Synth4K-4}}
& \multicolumn{2}{c}{\textbf{Synth4K-5}} \\
\cmidrule(lr){3-4}
\cmidrule(lr){5-6}
\cmidrule(lr){7-8}
\cmidrule(lr){9-10}
\cmidrule(lr){11-12}
& & AbsRel $\downarrow$ & $\delta_1\uparrow$
& AbsRel $\downarrow$ & $\delta_1\uparrow$
& AbsRel $\downarrow$ & $\delta_1\uparrow$
& AbsRel $\downarrow$ & $\delta_1\uparrow$
& AbsRel $\downarrow$ & $\delta_1\uparrow$ \\
\midrule
\multirow{9}{*}{\rotatebox{90}{\textbf{Overall}}}
& DA3 & 38.3 & 45.7 & 29.7 & 51.2 & 32.2 & 53.2 & 23.7 & 63.9 & 23.8 & 58.9 \\
& MoGe-2 & 22.1 & 67.1 & 16.6 & 76.6 & 18.0 & 79.3 & 14.1 & 82.9 & 11.0 & 85.9 \\
& PPD & 15.9 & 79.4 & 11.7 & 87.3 & 12.7 & 87.7 & 11.9 & 88.2 & 9.1 & 93.1 \\
& PointDiT & 44.2 & 42.6 & 40.4 & 40.2 & 47.7 & 36.6 & 29.6 & 55.6 & 39.1 & 38.4 \\
& URGT & 15.0 & 81.1 & 10.6 & 87.5 & 15.3 & 84.9 & \underline{7.4} & \underline{94.0} & 7.6 & 92.6 \\
& InfiniDepth & 31.3 & 59.0 & 21.7 & 69.6 & 33.0 & 56.7 & 17.5 & 80.8 & 29.9 & 65.3 \\
& InfiniDepth$^\dagger$ & \textbf{11.6} & \textbf{86.8} & \underline{9.4} & \underline{88.2} & 12.2 & 85.2 & \textbf{7.1} & \textbf{94.8} & \textbf{5.4} & \underline{95.6} \\
\cmidrule(l){2-12}
& \cellcolor{black!5}\textbf{Ours (1024)}
& 15.0 & 80.3 & 11.0 & 87.5 & \underline{11.3} & \underline{87.8}
& 9.5 & 91.4 & 8.9 & 92.2 \\
& \cellcolor{black!5}\textbf{Ours (MR)}
& \underline{12.3} & \underline{85.3} & \textbf{8.4} & \textbf{92.4}
& \textbf{8.5} & \textbf{92.9} & 7.7 & 93.6
& \underline{6.7} & \textbf{95.7} \\
\midrule
\multirow{9}{*}{\rotatebox{90}{\textbf{High-frequency}}}
& DA3 & 45.6 & 41.8 & 36.4 & 45.2 & 44.8 & 44.1 & 27.9 & 58.0 & 27.8 & 56.6 \\
& MoGe-2 & 29.4 & 58.0 & 24.7 & 62.7 & 31.5 & 61.9 & 19.9 & 72.1 & \underline{15.5} & 78.0 \\
& PPD & 29.4 & 62.3 & 24.5 & \underline{65.0} & 31.1 & \underline{65.0} & 20.3 & 73.5 & 18.4 & 75.6 \\
& PointDiT & 49.0 & 42.1 & 38.8 & 42.9 & 53.0 & 39.0 & 34.1 & 49.6 & 32.7 & 47.5 \\
& URGT & 30.0 & 62.3 & 28.3 & 63.1 & 44.6 & 61.7 & 20.4 & \underline{76.9} & 19.6 & 74.4 \\
& InfiniDepth & 51.1 & 44.2 & 36.9 & 49.6 & 56.4 & 39.3 & 36.2 & 64.4 & 49.2 & 46.1 \\
& InfiniDepth$^\dagger$ & 32.1 & \underline{67.3} & 28.2 & 64.7 & 34.9 & 63.5 & 28.2 & 76.7 & 19.3 & \underline{78.8} \\
\cmidrule(l){2-12}
& \cellcolor{black!5}\textbf{Ours (1024)}
& \underline{27.4} & 62.6 & \underline{24.3} & 64.7
& \underline{28.7} & 63.6 & \underline{19.4} & 74.7 & 17.2 & 74.7 \\
& \cellcolor{black!5}\textbf{Ours (MR)}
& \textbf{22.8} & \textbf{69.0} & \textbf{21.3} & \textbf{71.0}
& \textbf{24.0} & \textbf{70.7} & \textbf{16.7} & \textbf{78.8}
& \textbf{13.5} & \textbf{81.8} \\
\bottomrule
\end{tabular*}
\end{table}

\begin{table}[t]
\centering
\small
\setlength{\tabcolsep}{2.5pt}
\renewcommand{\arraystretch}{1.08}
\setlength{\belowcaptionskip}{6pt}
\vspace{-3mm}
\caption{\textbf{Boundary F1$\uparrow$ (radius = 1) results on Synth4K.} Our multi-resolution model achieves the highest scores across all five subsets, demonstrating excellent depth boundary quality.}
\label{tab:comparison-boundary}

\begin{tabular*}{\linewidth}{@{\extracolsep{\fill}}l*{5}{c}@{}}
\toprule
\textbf{Method}
& \textbf{Synth4K-1}
& \textbf{Synth4K-2}
& \textbf{Synth4K-3}
& \textbf{Synth4K-4}
& \textbf{Synth4K-5} \\
\midrule
DA3 & 8.7 & 10.8 & 9.2 & 11.1 & 14.1 \\
MoGe-2 & 15.7 & 13.1 & 13.2 & 14.9 & 18.2 \\
PPD & 15.0 & 18.4 & 15.8 & 18.7 & 16.1 \\
PointDiT & 20.7 & 21.8 & 17.4 & 22.1 & 20.1 \\
URGT & 20.9 & 16.0 & 15.0 & 18.2 & 21.4 \\
InfiniDepth & 15.7 & 20.0 & 13.1 & \underline{23.6} & 21.7 \\
InfiniDepth$^\dagger$ & 11.2 & 7.1 & 8.3 & 13.3 & 11.0 \\
\midrule
\cellcolor{black!5}\textbf{Ours (1024)}
& \underline{22.5} & \underline{22.2} & \underline{18.5}
& 22.5 & \underline{24.9} \\
\cellcolor{black!5}\textbf{Ours (MR)}
& \textbf{31.1} & \textbf{29.5} & \textbf{25.7}
& \textbf{30.4} & \textbf{32.5} \\
\bottomrule
\end{tabular*}
\vspace{-6mm}
\end{table}

\textbf{Quantitative results.}
Table~\ref{tab:comparison-standard} reports depth accuracy on five real-world benchmarks.
\methodname{} achieves competitive performance across these datasets, with both variants outperforming the comparison methods in AbsRel and $\delta_1$ on ETH3D and DIODE.
Ours (MR) maintains accuracy comparable to Ours (1024), indicating that mixed-resolution continuation largely preserves performance.
On Synth4K, evaluated at native 4K resolution, Ours (MR) achieves strong overall accuracy and leads all comparison methods in both HF-masked metrics and boundary F1 across all five subsets (Tables~\ref{tab:comparison-highres} and~\ref{tab:comparison-boundary}).
Meanwhile, the Ours (1024) model, which is trained at a lower resolution like most comparison methods, remains competitive in overall accuracy and fine-detail recovery at 4K.

\textbf{Efficiency analysis.}
Table~\ref{tab:efficiency} compares inference speed at two resolutions.
Our method is the fastest among the evaluated methods at 4K.
At $1024\times1024$, it is faster than all evaluated generative baselines while remaining competitive with discriminative methods.

\subsection{Ablation Studies}

\textbf{Decoding across different latent priors.} To evaluate the effectiveness of pixel diffusion decoding across different latent priors, we compare Lotus-2 and two Flux.2-depth variants trained with latent- and pixel-space supervision, respectively.
For each prior, we evaluate its standalone predictions and predictions obtained with an adapted PiD decoder.
As shown in Table~\ref{tab:ablation_priors}, PiD reduces overall AbsRel across all four datasets and improves both HF metrics on Synth4K-1/2 for all three priors, demonstrating its effectiveness in recovering fine-grained depth across different latent representations.
Among these priors, pixel-supervised Flux.2-depth provides the lowest overall AbsRel on all four datasets and the best HF metrics on both Synth4K subsets after pixel decoding.

\begin{table}[t]
\centering
\vspace{-3mm}
\caption{Pixel diffusion decoding across different latent priors. Flux.2-Depth denotes our depth-adapted latent diffusion model.
$\mathrm{AbsRel}^{H}$ and $\delta_1^{H}$ denote AbsRel and
$\delta_1$ evaluated in high-frequency masked regions, respectively.
}
\vspace{1.5mm}
\label{tab:ablation_priors}

\begingroup
\setlength{\tabcolsep}{1.5pt}
\renewcommand{\arraystretch}{1.08}
\small

\begin{adjustbox}{max width=\linewidth}
\begin{tabular}{@{}l*{12}{c}@{}}
\toprule
\multirow{2}{*}{\textbf{Configuration}}
& \multicolumn{2}{c}{\textbf{ETH3D}}
& \multicolumn{2}{c}{\textbf{DIODE}}
& \multicolumn{4}{c}{\textbf{Synth4K-1}}
& \multicolumn{4}{c}{\textbf{Synth4K-2}} \\
\cmidrule(lr){2-3}
\cmidrule(lr){4-5}
\cmidrule(lr){6-9}
\cmidrule(lr){10-13}
& AbsRel$\downarrow$ & $\delta_1\uparrow$
& AbsRel$\downarrow$ & $\delta_1\uparrow$
& AbsRel$\downarrow$ & $\delta_1\uparrow$
& AbsRel$^{H}\downarrow$ & $\delta_1^{H}\uparrow$
& AbsRel$\downarrow$ & $\delta_1\uparrow$
& AbsRel$^{H}\downarrow$ & $\delta_1^{H}\uparrow$ \\
\midrule
Lotus-2
& 5.1 & 97.5 & 6.1 & 95.2
& 28.0 & 62.7 & 47.0 & 45.1
& 24.7 & 68.0 & 46.1 & 40.5 \\

\quad + Pixel diffusion
& 4.1 & 98.3 & 4.8 & 96.9
& 17.9 & 74.8 & 31.3 & 57.4
& 13.2 & 83.7 & 25.8 & 60.1 \\

\addlinespace[3pt]

Flux.2-Depth (latent loss)
& 4.5 & 98.8 & 5.6 & 96.4
& 22.7 & 65.1 & 31.4 & 53.8
& 17.7 & 74.4 & 31.6 & 51.9 \\

\quad + Pixel diffusion
& 3.3 & 99.1 & 4.2 & \textbf{98.0}
& 15.9 & 78.8 & 29.4 & 59.8
& 12.6 & 84.9 & 26.6 & 60.9 \\

\addlinespace[3pt]

Flux.2-Depth (pixel loss)
& 3.6 & \textbf{99.2} & 4.6 & 97.7
& 19.9 & 70.7 & 29.1 & 56.9
& 16.6 & 77.8 & 29.4 & 52.4 \\

\quad + Pixel diffusion
& \textbf{3.1} & \textbf{99.2}
& \textbf{3.9} & 97.9
& \textbf{15.0} & \textbf{80.3}
& \textbf{27.4} & \textbf{62.6}
& \textbf{11.0} & \textbf{87.5}
& \textbf{24.3} & \textbf{64.7} \\
\bottomrule
\end{tabular}
\end{adjustbox}
\endgroup
\end{table}

\begin{table}[t]
\centering
\vspace{-3mm}
\caption{
\textbf{Effect of patch size in pixel diffusion.} We use a patch size of 16 for the best accuracy. Despite reduced boundary detail (F1), a patch size of 32 retains competitive overall depth accuracy while providing a 3.1× speedup at $3840\times2112$ resolution.
}
\vspace{1.5mm}
\label{tab:ablation_patch}

\begingroup
\setlength{\tabcolsep}{1.5mm}
\renewcommand{\arraystretch}{1.08}
\small

\begin{adjustbox}{max width=\linewidth}
\begin{tabular}{@{}c*{11}{c}@{}}
\toprule
\multirow{2}{*}{\textbf{Patch}}
& \multicolumn{2}{c}{\textbf{ETH3D}}
& \multicolumn{2}{c}{\textbf{DIODE}}
& \multicolumn{5}{c}{\textbf{Synth4K-1}}
& \multicolumn{2}{c}{\textbf{Efficiency: Time (ms)}$\downarrow$} \\
\cmidrule(lr){2-3}
\cmidrule(lr){4-5}
\cmidrule(lr){6-10}
\cmidrule(lr){11-12}
& AbsRel$\downarrow$ & $\delta_1\uparrow$
& AbsRel$\downarrow$ & $\delta_1\uparrow$
& AbsRel$\downarrow$ & $\delta_1\uparrow$
& F1$\uparrow$
& AbsRel$^{H}\downarrow$ & $\delta_1^{H}\uparrow$
& $1024^2$
& $3840\times2112$ \\
\midrule
16
& 3.1 & 99.1
& 3.9 & 97.9
& 12.3 & 85.3 & 31.1 & 22.8 & 69.0
& 204 & 1,439 \\

32
& 3.3 & 98.9
& 4.0 & 97.8
& 12.7 & 85.1 & 17.6 & 25.4 & 66.8
& 172 & 468 \\

64
& 4.1 & 98.2
& 4.6 & 97.2
& 13.1 & 84.4 & 9.2 & 28.1 & 63.7
& 170 & 362 \\
\bottomrule
\end{tabular}
\end{adjustbox}
\endgroup
\end{table}

\begin{figure}[!t]
  \centering
  \vspace{1mm}

  \begin{minipage}[t]{0.55\linewidth}
    \vspace{0pt}
    \centering

    \includegraphics[width=\linewidth]{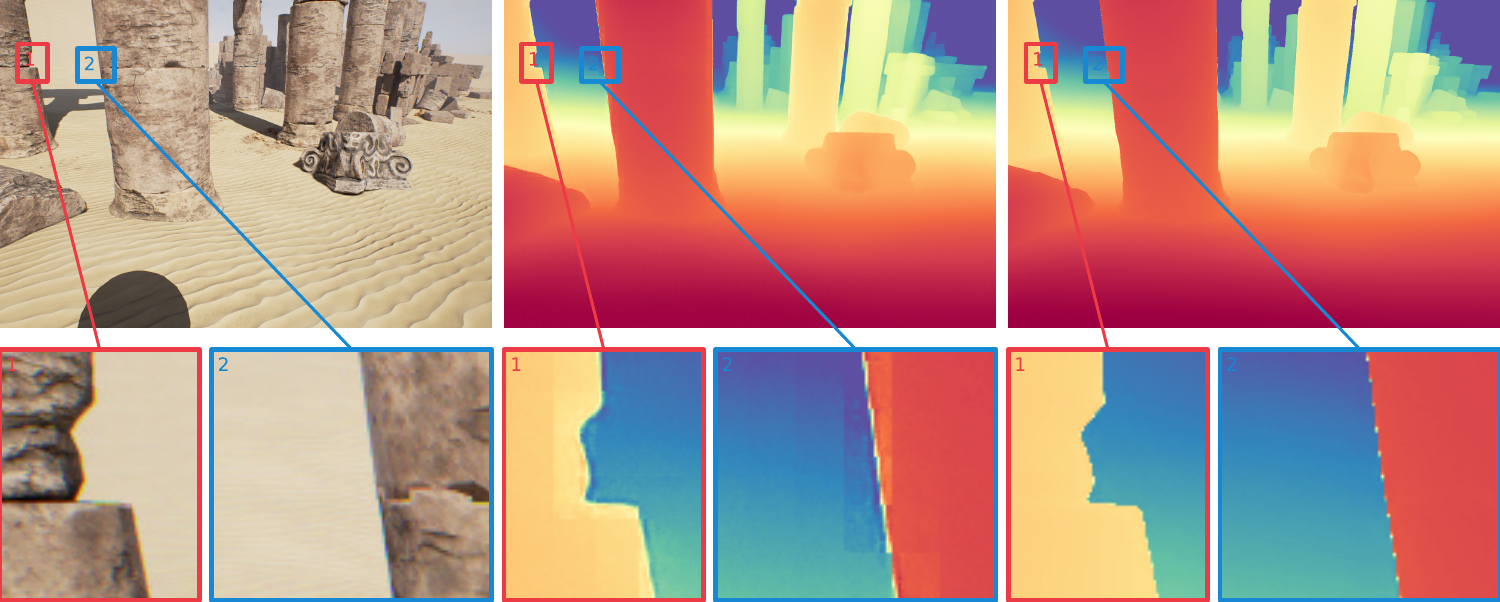}

    \par\vspace{2.2pt}
    \noindent
    \makebox[0.3333\linewidth][c]{RGB}\makebox[0.3333\linewidth][c]{w/o module}\makebox[0.3333\linewidth][c]{w/ module}\par

    \vspace{-1mm}
    \captionof{figure}{\textbf{Ablation study of the proposed continuity-promoting module.}
      The module effectively mitigates discontinuity artifacts between patches.
    }
    \label{fig:ablation_continuity}
  \end{minipage}\hfill\begin{minipage}[t]{0.42\linewidth}
  \vspace{0pt}
  \centering
  \vspace{-3.5mm}

  \captionof{table}{\textbf{Runtime comparison (ms$\downarrow$)}
    at $1024\times1024$ and $3840\times2160$.
  }
  \label{tab:efficiency}
  \vspace{1mm}

  \small
  \setlength{\tabcolsep}{3pt}
  \renewcommand{\arraystretch}{1.08}

  \begin{tabular*}{\linewidth}{@{\extracolsep{\fill}}cccc@{}}
    \toprule
    & Method & $1024\times1024$ & $3840\times2160$ \\
    \midrule

    \multirow{4}{*}{\rotatebox[origin=c]{90}{\scriptsize\textit{Discriminative}}}
      & DA3         & 105   & 1,784   \\
      & MoGe-2      & 82    & 1,637   \\
      & URGT        & -     & 24,306  \\
      & InfiniDepth & 142   & 1,740   \\

    \midrule

    \multirow{5}{*}{\rotatebox[origin=c]{90}{\scriptsize\textit{Generative}}}
      & Marigold    & 614   & 13,769  \\
      & Lotus-2     & 7,949 & 106,314 \\
      & PPD         & 899   & 27,577  \\
      & PointDiT    & 221   & 5,095   \\
      & \textbf{Ours}
                    & \textbf{204}
                            & \textbf{1,477} \\

    \bottomrule
  \end{tabular*}
\end{minipage}
\vspace{-4mm}
\end{figure}
    
\textbf{Effectiveness of continuity-promoting module.}
We compare the pixel decoder with and without the continuity-promoting module to assess its effect on patch-induced artifacts.
As shown in Figure~\ref{fig:ablation_continuity}, the variant without the module exhibits discontinuities at patch-size intervals, especially near abrupt depth transitions.
The module suppresses these artifacts through spatial feature mixing.

\textbf{Patch size in pixel diffusion.} Larger patches reduce the number of tokens and accelerate inference, but compromise depth accuracy.
We study this trade-off with patch sizes of 16, 32, and 64.
As shown in Table~\ref{tab:ablation_patch}, increasing the patch size degrades prediction quality, with boundary and HF metrics showing a greater loss of detail than the overall depth metrics suggest.
Notably, P32 retains similar overall accuracy to P16 while providing a $3.1\times$ speedup at $3840\times2112$ resolution, with degradation concentrated in fine-detail quality.
While we adopt P16 to prioritize fine-grained geometry, P32 offers an efficient alternative for applications with less requirements for depth details.

\section{Conclusion}
We presented \methodname{}, a framework that combines depth-adapted latent diffusion with pixel diffusion for efficient fine-grained depth estimation.
The latent branch predicts coarse depth latents from downsampled RGB inputs, while the pixel branch generates depth directly at the target resolution conditioned on these latents.
This design retains learned geometric priors while reducing detail loss from latent-space modeling and VAE reconstruction.
It also avoids the substantial overhead of VAE decoding at high resolutions.
We train two variants: one at approximately $1024^2$ pixels and a mixed-resolution extension supporting prediction up to 4K ($2160\times3840$).
Experiments on five commonly used depth estimation datasets and Synth4K demonstrate strong overall accuracy and better preservation of fine structures and object boundaries.

\bibliography{iclr2027_conference}

@article{lipman2022flow,
  title={Flow matching for generative modeling},
  author={Lipman, Yaron and Chen, Ricky TQ and Ben-Hamu, Heli and Nickel, Maximilian and Le, Matt},
  journal={arXiv preprint arXiv:2210.02747},
  year={2022}
}

@inproceedings{ke2024repurposing,
  title={Repurposing diffusion-based image generators for monocular depth estimation},
  author={Ke, Bingxin and Obukhov, Anton and Huang, Shengyu and Metzger, Nando and Daudt, Rodrigo Caye and Schindler, Konrad},
  booktitle={2024 IEEE/CVF Conference on Computer Vision and Pattern Recognition (CVPR)},
  pages={9492--9502},
  year={2024},
  organization={IEEE}
}

@inproceedings{zhang2026dvd,
  title={Dvd: Deterministic video depth estimation with generative priors},
  author={Zhang, Hongfei and Chen, Harold Haodong and Liao, Chenfei and He, Jing and Zhang, Zixin and Li, Haodong and Liang, Yihao and Chen, Kanghao and Ren, Bin and Zheng, Xu and others},
  booktitle={ICML 2026 Workshop on Foundations of Deep Generative Models: Understanding Memorization, Generalization, and Reasoning},
  year={2026}
}

@article{he2025lotus2,
  title={Lotus-2: Advancing geometric dense prediction with powerful image generative model},
  author={He, Jing and Li, Haodong and Sheng, Mingzhi and Chen, Ying-Cong},
  journal={arXiv preprint arXiv:2512.01030},
  year={2025}
}

@inproceedings{he2025lotus,
  title={Lotus: Diffusion-based visual foundation model for high-quality dense prediction},
  author={He, Jing and Li, Haodong and Yin, Wei and Liang, Yixun and Li, Leheng and Zhou, Kaiqiang and Zhang, Hongbo and Liu, Bingbing and Chen, YingCong},
  booktitle={International Conference on Learning Representations},
  volume={2025},
  pages={89454--89467},
  year={2025}
}

@article{lu2026pid,
  title={PiD: Fast and High-Resolution Latent Decoding with Pixel Diffusion},
  author={Lu, Yifan and Wu, Qi and Wu, Jay Zhangjie and Wang, Zian and Ling, Huan and Fidler, Sanja and Ren, Xuanchi},
  journal={arXiv preprint arXiv:2605.23902},
  year={2026}
}

@inproceedings{frans2025one,
  title={One step diffusion via shortcut models},
  author={Frans, Kevin and Hafner, Danijar and Levine, Sergey and Abbeel, Pieter},
  booktitle={International Conference on Learning Representations},
  volume={2025},
  pages={34668--34684},
  year={2025}
}

@article{yang2024depth,
  title={Depth anything v2},
  author={Yang, Lihe and Kang, Bingyi and Huang, Zilong and Zhao, Zhen and Xu, Xiaogang and Feng, Jiashi and Zhao, Hengshuang},
  journal={Advances in neural information processing systems},
  volume={37},
  pages={21875--21911},
  year={2024}
}

@inproceedings{yin2019enforcing,
  title={Enforcing geometric constraints of virtual normal for depth prediction},
  author={Yin, Wei and Liu, Yifan and Shen, Chunhua and Yan, Youliang},
  booktitle={2019 IEEE/CVF International Conference on Computer Vision (ICCV)},
  pages={5683--5692},
  year={2019},
  organization={IEEE}
}

@article{wang2026vggt,
  title={VGGT-{$\Omega$}},
  author={Wang, Jianyuan and Chen, Minghao and Zhang, Shangzhan and Karaev, Nikita and Sch{\"o}nberger, Johannes and Labatut, Patrick and Bojanowski, Piotr and Novotny, David and Vedaldi, Andrea and Rupprecht, Christian},
  journal={arXiv preprint arXiv:2605.15195},
  year={2026}
}

@inproceedings{roberts2021hypersim,
  title={Hypersim: A photorealistic synthetic dataset for holistic indoor scene understanding},
  author={Roberts, Mike and Ramapuram, Jason and Ranjan, Anurag and Kumar, Atulit and Bautista, Miguel Angel and Paczan, Nathan and Webb, Russ and Susskind, Joshua M},
  booktitle={2021 IEEE/CVF International Conference on Computer Vision (ICCV)},
  pages={10892--10902},
  year={2021},
  organization={IEEE}
}

@article{cabon2020virtual,
  title={Virtual kitti 2},
  author={Cabon, Yohann and Murray, Naila and Humenberger, Martin},
  journal={arXiv preprint arXiv:2001.10773},
  year={2020}
}

@inproceedings{wang2020tartanair,
  title={Tartanair: A dataset to push the limits of visual slam},
  author={Wang, Wenshan and Zhu, Delong and Wang, Xiangwei and Hu, Yaoyu and Qiu, Yuheng and Wang, Chen and Hu, Yafei and Kapoor, Ashish and Scherer, Sebastian},
  booktitle={2020 IEEE/RSJ International Conference on Intelligent Robots and Systems (IROS)},
  pages={4909--4916},
  year={2020},
  organization={IEEE}
}

@inproceedings{wang2021irs,
  title={Irs: A large naturalistic indoor robotics stereo dataset to train deep models for disparity and surface normal estimation},
  author={Wang, Qiang and Zheng, Shizhen and Yan, Qingsong and Deng, Fei and Zhao, Kaiyong and Chu, Xiaowen},
  booktitle={2021 IEEE International Conference on Multimedia and Expo (ICME)},
  pages={1--6},
  year={2021},
  organization={IEEE}
}

@inproceedings{le2021eden,
  title={Eden: Multimodal synthetic dataset of enclosed garden scenes},
  author={Le, Hoang-An and Mensink, Thomas and Das, Partha and Karaoglu, Sezer and Gevers, Theo},
  booktitle={2021 IEEE Winter Conference on Applications of Computer Vision (WACV)},
  pages={1578--1588},
  year={2021},
  organization={IEEE}
}

@article{lingbot-depth2026,
  title={Masked Depth Modeling for Spatial Perception},
  author={Tan, Bin and Sun, Changjiang and Qin, Xiage and Adai, Hanat and Fu, Zelin and Zhou, Tianxiang and Zhang, Han and Xu, Yinghao and Zhu, Xing and Shen, Yujun and Xue, Nan},
  journal={arXiv preprint arXiv:2601.17895},
  year={2026}
}

@inproceedings{zhang2018unrealstereo,
  title={Unrealstereo: Controlling hazardous factors to analyze stereo vision},
  author={Zhang, Yi and Qiu, Weichao and Chen, Qi and Hu, Xiaolin and Yuille, Alan},
  booktitle={2018 International Conference on 3D Vision (3DV)},
  pages={228--237},
  year={2018},
  organization={IEEE}
}

@article{gomez2025all,
  title={All for one, and one for all: Urbansyn dataset, the third musketeer of synthetic driving scenes},
  author={G{\'o}mez, Jose L and Silva, Manuel and Seoane, Antonio and Borr{\'a}s, Agnes and Noriega, Mario and Ros, Germ{\'a}n and Iglesias-Guitian, Jose A and L{\'o}pez, Antonio M},
  journal={Neurocomputing},
  volume={637},
  pages={130038},
  year={2025},
  publisher={Elsevier}
}

@inproceedings{mehl2023spring,
  title={Spring: A high-resolution high-detail dataset and benchmark for scene flow, optical flow and stereo},
  author={Mehl, Lukas and Schmalfuss, Jenny and Jahedi, Azin and Nalivayko, Yaroslava and Bruhn, Andr{\'e}s},
  booktitle={2023 IEEE/CVF Conference on Computer Vision and Pattern Recognition (CVPR)},
  pages={4981--4991},
  year={2023},
  organization={IEEE}
}

@inproceedings{silberman2012indoor,
  title={Indoor segmentation and support inference from rgbd images},
  author={Silberman, Nathan and Hoiem, Derek and Kohli, Pushmeet and Fergus, Rob},
  booktitle={European conference on computer vision},
  pages={746--760},
  year={2012},
  organization={Springer}
}

@inproceedings{geiger2012we,
  title={Are we ready for autonomous driving? the kitti vision benchmark suite},
  author={Geiger, Andreas and Lenz, Philip and Urtasun, Raquel},
  booktitle={2012 IEEE conference on computer vision and pattern recognition},
  pages={3354--3361},
  year={2012},
  organization={IEEE}
}

@inproceedings{schops2017multi,
  title={A multi-view stereo benchmark with high-resolution images and multi-camera videos},
  author={Schops, Thomas and Schonberger, Johannes L and Galliani, Silvano and Sattler, Torsten and Schindler, Konrad and Pollefeys, Marc and Geiger, Andreas},
  booktitle={Proceedings of the IEEE conference on computer vision and pattern recognition},
  pages={3260--3269},
  year={2017}
}

@inproceedings{dai2017scannet,
  title={Scannet: Richly-annotated 3d reconstructions of indoor scenes},
  author={Dai, Angela and Chang, Angel X and Savva, Manolis and Halber, Maciej and Funkhouser, Thomas and Nie{\ss}ner, Matthias},
  booktitle={2017 IEEE conference on computer vision and pattern recognition (CVPR)},
  pages={2432--2443},
  year={2017},
  organization={IEEE}
}

@article{vasiljevic2019diode,
  title={Diode: A dense indoor and outdoor depth dataset},
  author={Vasiljevic, Igor and Kolkin, Nick and Zhang, Shanyi and Luo, Ruotian and Wang, Haochen and Dai, Falcon Z and Daniele, Andrea F and Mostajabi, Mohammadreza and Basart, Steven and Walter, Matthew R and others},
  journal={arXiv preprint arXiv:1908.00463},
  year={2019}
}

@article{yu2026infinidepth,
  title={Infinidepth: Arbitrary-resolution and fine-grained depth estimation with neural implicit fields},
  author={Yu, Hao and Lin, Haotong and Wang, Jiawei and Li, Jiaxin and Wang, Yida and Zhang, Xueyang and Wang, Yue and Zhou, Xiaowei and Hu, Ruizhen and Peng, Sida},
  journal={arXiv preprint arXiv:2601.03252},
  year={2026}
}

@misc{flux-2-2025,
    author={Black Forest Labs},
    title={{FLUX.2: Frontier Visual Intelligence}},
    year={2025},
    howpublished={\url{https://bfl.ai/blog/flux-2}},
}

@inproceedings{bochkovskiy2025depth,
  title={Depth pro: Sharp monocular metric depth in less than a second},
  author={Bochkovskiy, Alexey and Delaunoy, Ama{\"e}l and Germain, Hugo and Santos, Marcel and Zhou, Yichao and Richter, Stephan and Koltun, Vladlen},
  booktitle={International Conference on Learning Representations},
  volume={2025},
  pages={75602--75637},
  year={2025}
}

@article{lin2025depth,
  title={Depth anything 3: Recovering the visual space from any views},
  author={Lin, Haotong and Chen, Sili and Liew, Junhao and Chen, Donny Y and Li, Zhenyu and Shi, Guang and Feng, Jiashi and Kang, Bingyi},
  journal={arXiv preprint arXiv:2511.10647},
  year={2025}
}

@article{ranftl2020towards,
  title={Towards robust monocular depth estimation: Mixing datasets for zero-shot cross-dataset transfer},
  author={Ranftl, Ren{\'e} and Lasinger, Katrin and Hafner, David and Schindler, Konrad and Koltun, Vladlen},
  journal={IEEE transactions on pattern analysis and machine intelligence},
  volume={44},
  number={3},
  pages={1623--1637},
  year={2020},
  publisher={IEEE}
}

@article{xu2026pixel,
  title={Pixel-perfect depth with semantics-prompted diffusion transformers},
  author={Xu, Gangwei and Lin, Haotong and Luo, Hongcheng and Wang, Xianqi and Yao, Jingfeng and Zhu, Lianghui and Pu, Yuechuan and Chi\_, Cheng and Sun, Haiyang and Wang, Bing and others},
  journal={Advances in Neural Information Processing Systems},
  volume={38},
  pages={174731--174755},
  year={2026}
}

@article{wang2026moge,
  title={Moge-2: Accurate monocular geometry with metric scale and sharp details},
  author={Wang, Ruicheng and Xu, Sicheng and Dong, Yue and Deng, Yu and Xiang, Jianfeng and Lv, Zelong and Sun, Guangzhong and Tong, Xin and Yang, Jiaolong},
  journal={Advances in Neural Information Processing Systems},
  volume={38},
  pages={35928--35959},
  year={2026}
}

@article{xu2026pointdit,
  title={PointDiT: Pixel-Space Diffusion for Monocular Geometry Estimation},
  author={Xu, Haofei and Wu, Rundi and Henzler, Philipp and Kalischek, Nikolai and Oechsle, Michael and Manhardt, Fabian and Pollefeys, Marc and Geiger, Andreas and Tombari, Federico and Niemeyer, Michael},
  journal={arXiv preprint arXiv:2607.02515},
  year={2026}
}

@inproceedings{li2026patchrefiner,
  title={Patchrefiner v2: Fast and lightweight real-domain high-resolution metric depth estimation},
  author={Li, Zhenyu and Cui, Wenqing and Bhat, Shariq and Wonka, Peter},
  booktitle={International Conference on Learning Representations},
  volume={2026},
  pages={117738--117753},
  year={2026}
}

@inproceedings{cui2026any,
  title={Any resolution any geometry: From multi-view to multi-patch},
  author={Cui, Wenqing and Li, Zhenyu and Lavreniuk, Mykola and Shi, Jian and Idoughi, Ramzi and Tang, Xiangjun and Wonka, Peter},
  booktitle={Proceedings of the IEEE/CVF Conference on Computer Vision and Pattern Recognition},
  pages={12576--12585},
  year={2026}
}

@article{2,
  title={Depth map prediction from a single image using a multi-scale deep network},
  author={Eigen, David and Puhrsch, Christian and Fergus, Rob},
  journal={Advances in neural information processing systems},
  volume={27},
  year={2014}
}

@inproceedings{3,
  title={Octocc: High-resolution 3d occupancy prediction with octree},
  author={Ouyang, Wenzhe and Song, Xiaolin and Feng, Bailan and Xu, Zenglin},
  booktitle={Proceedings of the AAAI Conference on Artificial Intelligence},
  volume={38},
  number={5},
  pages={4369--4377},
  year={2024}
}

@inproceedings{4,
  title={Mvdiffusion++: A dense high-resolution multi-view diffusion model for single or sparse-view 3d object reconstruction},
  author={Tang, Shitao and Chen, Jiacheng and Wang, Dilin and Tang, Chengzhou and Zhang, Fuyang and Fan, Yuchen and Chandra, Vikas and Furukawa, Yasutaka and Ranjan, Rakesh},
  booktitle={European Conference on Computer Vision},
  pages={175--191},
  year={2024},
  organization={Springer}
}

@article{yuan2025depthvla,
  title={Depthvla: Enhancing vision-language-action models with depth-aware spatial reasoning},
  author={Yuan, Tianyuan and Liu, Yicheng and Lu, Chenhao and Chen, Zhuoguang and Jiang, Tao and Zhao, Hang},
  journal={arXiv preprint arXiv:2510.13375},
  year={2025}
}

@inproceedings{litemvs,
  title={LiteMVS: Efficient Multi-View Stereo with Foundation Distillation and Expert Aggregation},
  author={Zhang, Tianbao and Liu, Zeyu and Wu, Shuyu and Li, Fanxing and Fan, Zhaoxin and Zou, Danping and others},
  booktitle={Proceedings of the IEEE/CVF Conference on Computer Vision and Pattern Recognition},
  pages={4520--4529},
  year={2026}
}

@inproceedings{2k,
  title={2K Retrofit: Entropy-Guided Efficient Sparse Refinement for High-Resolution 3D Geometry Prediction},
  author={Zhang, Tianbao and Liang, Zhenyu and Song, Zhenbo and Wang, Nana and Zhang, Xiaomei and Cai, Xudong and Zhu, Zheng and Wu, Kejian and Wang, Gang and Fan, Zhaoxin},
  booktitle={European Conference on Computer Vision},
  pages={150--169},
  year={2026},
  organization={Springer}
}

@article{vds,
  title={Vds-nav: Volumetric depth-based safe navigation for aerial robots--bridging the sim-to-real gap},
  author={Redder, Adrian and Pham, Huy Xuan and Sarabakha, Andriy and Kayacan, Erdal and others},
  journal={IEEE Robotics and Automation Letters},
  year={2025},
  publisher={IEEE}
}

@inproceedings{flownav,
  title={Flownav: Combining flow matching and depth priors for efficient navigation},
  author={Gode, Samiran and Nayak, Abhijeet and Oliveira, D{\'e}bora NP and Krawez, Michael and Schmid, Cordelia and Burgard, Wolfram},
  booktitle={2025 IEEE/RSJ International Conference on Intelligent Robots and Systems (IROS)},
  pages={17762--17768},
  year={2025},
  organization={IEEE}
}

@inproceedings{patchfusion,
  title={Patchfusion: An end-to-end tile-based framework for high-resolution monocular metric depth estimation},
  author={Li, Zhenyu and Bhat, Shariq Farooq and Wonka, Peter},
  booktitle={2024 IEEE/CVF Conference on Computer Vision and Pattern Recognition (CVPR)},
  pages={10016--10025},
  year={2024},
  organization={IEEE}
}

@inproceedings{marigold,
  title={Repurposing diffusion-based image generators for monocular depth estimation},
  author={Ke, Bingxin and Obukhov, Anton and Huang, Shengyu and Metzger, Nando and Daudt, Rodrigo Caye and Schindler, Konrad},
  booktitle={2024 IEEE/CVF Conference on Computer Vision and Pattern Recognition (CVPR)},
  pages={9492--9502},
  year={2024},
  organization={IEEE}
}

@inproceedings{geowizard,
  title={Geowizard: Unleashing the diffusion priors for 3d geometry estimation from a single image},
  author={Fu, Xiao and Yin, Wei and Hu, Mu and Wang, Kaixuan and Ma, Yuexin and Tan, Ping and Shen, Shaojie and Lin, Dahua and Long, Xiaoxiao},
  booktitle={European Conference on Computer Vision},
  pages={241--258},
  year={2024},
  organization={Springer}
}

@inproceedings{hu2025depthcrafter,
  title={Depthcrafter: Generating consistent long depth sequences for open-world videos},
  author={Hu, Wenbo and Gao, Xiangjun and Li, Xiaoyu and Zhao, Sijie and Cun, Xiaodong and Zhang, Yong and Quan, Long and Shan, Ying},
  booktitle={2025 IEEE/CVF Conference on Computer Vision and Pattern Recognition (CVPR)},
  pages={2005--2015},
  year={2025},
  organization={IEEE}
}

@inproceedings{ldm,
  title={High-resolution image synthesis with latent diffusion models},
  author={Rombach, Robin and Blattmann, Andreas and Lorenz, Dominik and Esser, Patrick and Ommer, Bj{\"o}rn},
  booktitle={2022 IEEE/CVF conference on computer vision and pattern recognition (CVPR)},
  pages={10674--10685},
  year={2022},
  organization={ieee}
}

@article{magicvideo,
  title={Magicvideo: Efficient video generation with latent diffusion models},
  author={Zhou, Daquan and Wang, Weimin and Yan, Hanshu and Lv, Weiwei and Zhu, Yizhe and Feng, Jiashi},
  journal={arXiv preprint arXiv:2211.11018},
  year={2022}
}

@inproceedings{jit,
  title={Back to basics: Let denoising generative models denoise},
  author={Li, Tianhong and He, Kaiming},
  booktitle={Proceedings of the IEEE/CVF Conference on Computer Vision and Pattern Recognition},
  pages={36115--36125},
  year={2026}
}

@inproceedings{pixeldit,
  title={Pixeldit: Pixel diffusion transformers for image generation},
  author={Yu, Yongsheng and Xiong, Wei and Nie, Weili and Sheng, Yichen and Liu, Shiqiu and Luo, Jiebo},
  booktitle={Proceedings of the IEEE/CVF Conference on Computer Vision and Pattern Recognition},
  pages={14273--14282},
  year={2026}
}

@inproceedings{ranftl2021vision,
  title={Vision transformers for dense prediction},
  author={Ranftl, Ren{\'e} and Bochkovskiy, Alexey and Koltun, Vladlen},
  booktitle={2021 IEEE/CVF International Conference on Computer Vision (ICCV)},
  pages={12159--12168},
  year={2021},
  organization={IEEE}
}

@article{ren2026anydepth,
  title={AnyDepth: Depth Estimation Made Easy},
  author={Ren, Zeyu and Zhang, Zeyu and Li, Wukai and Tang, Hao},
  year={2026}
}

@article{rawmantic2026data,
  title={Rawmantic 3D Training Field: From The Real World, For The Real World},
  author={Rawmantic Team},
  journal={Rawmantic},
  year={2026}
}

@misc{bfl2026flux2klein,
  author       = {{Black Forest Labs}},
  title        = {{FLUX.2 [klein]}: Towards Interactive Visual Intelligence},
  year         = {2026},
  howpublished = {\url{https://bfl.ai/blog/flux2-klein-towards-interactive-visual-intelligence}}
}
\bibliographystyle{iclr2027_conference}

\newpage
\appendix
\section{Architecture Details}

\subsection{Latent backbone: FLUX.2}

We use FLUX.2-klein-4B~\citep{flux-2-2025} as the latent backbone.
This pretrained rectified-flow transformer contains approximately four billion parameters, with five double-stream blocks followed by twenty single-stream blocks.
All blocks use a hidden dimension of 3072 and 24 attention heads.
The double-stream blocks maintain separate image and text projections and feed-forward networks, while jointly attending to tokens from both modalities.
The single-stream blocks process the concatenated sequence of image and text tokens with parallel attention and feed-forward branches.
The backbone uses query and key normalization, rotary positional embeddings, and timestep-conditioned feature modulation.
Its Qwen3-4B text encoder concatenates features from three intermediate layers into 7680-dimensional text tokens, which are projected to the transformer width.

The accompanying FLUX.2 autoencoder uses a convolutional encoder--decoder with residual blocks and a 32-channel latent representation at one-eighth of the image height and width.
Packing each $2\times2$ latent neighborhood into the channel dimension produces 128-channel features at a spatial stride of 16.
The encoder uses the posterior mean and normalizes the packed features using fixed pretrained channel statistics.
For an input image of size $H\times W$, the resulting latent representation has spatial dimensions $H/16\times W/16$ and 128 channels.
The transformer predicts a velocity in this same representation.

We initialize the latent predictor from the released distilled checkpoint and adapt its transformer blocks with rank-256 LoRA.
The encoded RGB image serves as the input state at $t^\star=0.5$, and the depth latent is obtained by the single flow update in Eq.~\ref{eq:coarse-depth}.
For text conditioning, we use the fixed prompt ``A dense grayscale monocular depth map of the scene, far surfaces brighter and near surfaces darker.''
Its embedding is computed once and cached.
The autoencoder remains frozen; its decoder is used for the coarse-depth supervision described in the main paper.
During pixel adaptation and final inference, the latent predictor remains frozen and operates at approximately $512^2$ input pixels, passing its predicted depth latent directly to the pixel diffusion decoder.

\subsection{Pixel backbone: PiD}

We use pretrained PiD~\citep{lu2026pid} as the pixel backbone, adopting its FLUX.2-compatible v1.5 architecture.
Built on PixelDiT~\citep{pixeldit}, PiD combines a patch transformer with a pixel pathway to jointly perform latent decoding and spatial upsampling.
The patch transformer processes $16\times16$ image patches using 14 MM-DiT blocks with hidden dimension 1536 and 24 attention heads, allowing image and text tokens to interact through joint attention.
The pixel pathway uses two PiT blocks with 16-dimensional pixel features, conditioned on the patch representations through adaptive normalization.
Each PiT block compresses the features within each patch for attention across patches, then expands them back for pixel updates.

A convolutional adapter maps the conditioning latent to features aligned with the image patch grid.
For FLUX.2 latents, the packed 128-channel representation is first unpacked into 32 channels and resized by nearest-neighbor interpolation to $H/p\times W/p$, where $p=16$.
Two $3\times3$ convolutions with an intermediate SiLU activation and four residual blocks extract features with 1024 channels.
These features are flattened into tokens and independently projected to the backbone width:
\begin{equation}
    \begin{aligned}
        \boldsymbol{z}_{\mathrm{grid}}
        &=
        \operatorname{Resize}\!\left(
            \operatorname{Unpack}_{2\times2}
            \!\left(\tilde{\boldsymbol{z}}_{\sigma_{\mathrm{lat}}}\right)
        \right),\\
        \boldsymbol{\ell}^{(k)}
        &=
        \operatorname{Linear}_{k}\!\left(
            \operatorname{Flatten}\!\left(
                \operatorname{ConvBlocks}(\boldsymbol{z}_{\mathrm{grid}})
            \right)
        \right),
        \qquad k=1,\ldots,7,
    \end{aligned}
    \label{eq:pid-latent-projection}
\end{equation}
where $\tilde{\boldsymbol{z}}_{\sigma_{\mathrm{lat}}}$ denotes the conditioning latent at noise level $\sigma_{\mathrm{lat}}$, and $\operatorname{ConvBlocks}$ denotes the convolutional feature extractor described above.
Each $\boldsymbol{\ell}^{(k)}\in\mathbb R^{N\times1536}$ contains one conditioning feature per image patch, with $N=HW/p^2$.

Original PiD trains with corrupted conditions
$\tilde{\boldsymbol{z}}_{\sigma_{\mathrm{lat}}}=(1-\sigma_{\mathrm{lat}})\boldsymbol{z}+\sigma_{\mathrm{lat}}\boldsymbol{\epsilon}$,
where $\boldsymbol{z}$ is the uncorrupted latent and $\boldsymbol{\epsilon}\sim\mathcal N(\boldsymbol{0},\boldsymbol{I})$, to support decoding from partially denoised latents.
A sigma-aware gate regulates latent injection before alternating MM-DiT blocks:
\begin{equation}
    \begin{aligned}
        \boldsymbol{g}^{(k)}
        &=
        \operatorname{sigmoid}\!\left(
            \operatorname{Linear}^{\mathrm{gate}}_{k}
            \!\left([\boldsymbol{s}^{(k)};\boldsymbol{\ell}^{(k)}]\right)
            -\alpha_k\sigma_{\mathrm{lat}}
        \right),\\
        \tilde{\boldsymbol{s}}^{(k)}
        &=\boldsymbol{s}^{(k)}+\boldsymbol{g}^{(k)}\odot\boldsymbol{\ell}^{(k)},
    \end{aligned}
    \label{eq:pid-latent-injection}
\end{equation}
where $\boldsymbol{s}^{(k)}$ denotes the image tokens before block $2k-1$, $[\cdot;\cdot]$ denotes channel concatenation, and $\alpha_k>0$ is learned.
In the v1.5 implementation, each gate produces one scalar per token, broadcast across channels.
The content term learns the injection strength, while the noise-dependent offset reduces it as $\sigma_{\mathrm{lat}}$ increases.
An additional projection and gate condition the patch features supplied to both PiT blocks.

For our task, we use the observed RGB image as the pixel input state and the coarse predictor's depth latent $\hat{\boldsymbol{z}}_{\mathrm{depth}}^{L}$ as the conditioning signal.
For simplicity, we treat this predicted latent as clean: no noise is added, and $\sigma_{\mathrm{lat}}=0$ throughout training and inference.
Thus, we retain content-dependent gating without noise-level modulation.
We freeze the coarse predictor and fine-tune PiD and its adapter for single-step depth prediction at $t^\star=0.5$, retaining a fixed text embedding cache.
The decoder directly predicts the clean sample (x-pred), with the continuity-promoting module inserted before the original output projection as described in the main paper.
Its three output channels are averaged to obtain the depth map.

\section{Dataset Details}
To improve training efficiency and reduce storage requirements, we filter and subsample the larger training datasets.
The aim is to retain informative samples while reducing low-quality supervision and redundant observations.

For Hypersim, VKITTI2, TartanAir, and IRS, we adopt training subsets based on the quality filtering of AnyDepth~\citep{ren2026anydepth}, which screens samples according to valid-depth coverage, depth-value distribution, and gradient continuity.
For EDEN, we use the smaller sample subset provided by the dataset authors.
For RobbyReal and Rawmantic3D, we observe substantial similarity between neighboring frames within a scene and therefore retain every tenth frame from each sequence.
For UnrealStereo4K, UrbanSyn, and Spring, we retain all available training samples because these datasets are relatively small and provide valuable supervision for high-resolution depth prediction.
Table~\ref{tab:training_data} summarizes the number of samples used from each dataset, totaling 535,562 image and depth pairs (approximately 536K).

\begin{table}[t]
    \centering
    \caption{\textbf{Training datasets and sample selection strategies.} We report depth resolutions ($H \times W$), selection criteria, and the number of selected RGB–depth pairs for each dataset.}
    \label{tab:training_data}
    \begin{tabular}{lclr}
        \toprule
        Dataset & Depth resolution & Selection & Samples \\
        \midrule
        Hypersim       & $768\times1024$                & AnyDepth filtering & 26,695 \\
        VKITTI2        & $375\times1242$                & AnyDepth filtering & 12,643 \\
        TartanAir      & $480\times640$                 & AnyDepth filtering & 186,693 \\
        IRS            & $540\times960$                 & AnyDepth filtering & 68,211 \\
        EDEN           & $480\times640$                 & Official subset    & 12,520 \\
        RobbyReal      & $480\times640$, $720\times1280$ & Every tenth frame  & 139,575 \\
        Rawmantic3D    & $1080\times1920$               & Every tenth frame  & 68,686 \\
        UnrealStereo4K & $2160\times3840$               & All samples        & 8,000 \\
        UrbanSyn       & $1024\times2048$               & All samples        & 7,539 \\
        Spring         & $2160\times3840$               & All samples        & 5,000 \\
        \midrule
        \textbf{Total} & & & \textbf{535,562} \\
        \bottomrule
    \end{tabular}
\end{table}

\section{Training Details}
\label{app:training}

\subsection{Data Augmentation}

We use only random horizontal flipping and gamma augmentation to improve robustness to scene orientation and photometric variations.
Both are applied with probability $0.5$, with gamma augmentation defined as $I'=I^\gamma$ for RGB intensities $I\in[0,1]$ and $\gamma\sim\mathcal U(0.8,1.2)$.
The high-resolution stream uses horizontal flipping only.

\subsection{Multi-Aspect-Ratio and Mixed-Resolution Training}

\textbf{Aspect-ratio bucketing.}
Excessive stretching or compression distorts object shapes and proportions, potentially weakening semantic and geometric cues for depth estimation.
At each resolution level, we therefore use six representative aspect ratios: $1\!:\!1$, $4\!:\!3$, $3\!:\!2$, $16\!:\!9$, $2\!:\!1$, and $10\!:\!3$, expressed as width to height.
Images are cropped to the selected aspect ratio before resizing to preserve object proportions.
During bucketed inference, images whose height exceeds their width are rotated by $90^\circ$, and the predicted depth is rotated back to the original orientation.
Table~\ref{tab:training_resolutions} lists the corresponding image dimensions.

\begin{table}[t]
  \centering
  \caption{\textbf{Training resolution buckets for our models.} Aspect ratios are given as $W:H$ and image dimensions as $H \times W$. All dimensions are divisible by 16 to accommodate the models' patch size.}
  \label{tab:training_resolutions}
  \small
  \setlength{\tabcolsep}{3pt}
  \begin{adjustbox}{width=\linewidth}
    \begin{tabular}{@{}cccccc@{}}
      \toprule
      & Latent branch & \multicolumn{4}{c}{Pixel decoder} \\
      \cmidrule(lr){2-2}\cmidrule(lr){3-6}
      Aspect ratio & 512-scale & 1024-scale & High-res. low & High-res. middle & High-res. high \\
      \midrule
      $1:1$  & $512\times512$ & $1024\times1024$ & $1536\times1536$ & $2048\times2048$ & $2160\times2160$ \\
      $4:3$  & $432\times576$ & $864\times1152$  & $1344\times1792$ & $1776\times2368$ & $2160\times2880$ \\
      $3:2$  & $416\times624$ & $832\times1248$  & $1248\times1872$ & $1664\times2496$ & $2144\times3216$ \\
      $16:9$ & $384\times688$ & $720\times1280$  & $1152\times2048$ & $1584\times2816$ & $2160\times3840$ \\
      $2:1$  & $368\times736$ & $720\times1440$  & $1088\times2176$ & $1456\times2912$ & $1920\times3840$ \\
      $10:3$ & $288\times960$ & $576\times1920$  & $864\times2880$  & $1104\times3680$ & $1152\times3840$ \\
      \bottomrule
    \end{tabular}
  \end{adjustbox}
\end{table}

\textbf{Mixed-resolution sampling.}
Coarse depth prediction operates at approximately $512^2$ pixels to provide scene-level depth structure.
The pixel diffusion decoder is first trained at approximately $1024^2$ pixels, allowing it to learn depth prediction at moderate computational cost.
We then continue training with mixed resolutions up to 4K ($2160\times3840$) to adapt the model to a wider range of output resolutions, while keeping the coarse predictor frozen at the 512-scale setting.
The high-resolution buckets comprise three levels (low, medium, and high) which are trained jointly with 1024-scale replay to mitigate forgetting.
The high-resolution stream uses only UnrealStereo4K, UrbanSyn, Spring, and Rawmantic3D to learn fine geometric details and depth boundaries.
The four levels (1024-scale and the three high-resolution levels) are sampled with equal probability, each accounting for $25\%$ of training batches in expectation.

\section{Additional Results}
\label{app:additional-results}

\subsection{More Qualitative Results}

\begin{figure}[t]
    \centering
    \includegraphics[width=\linewidth]{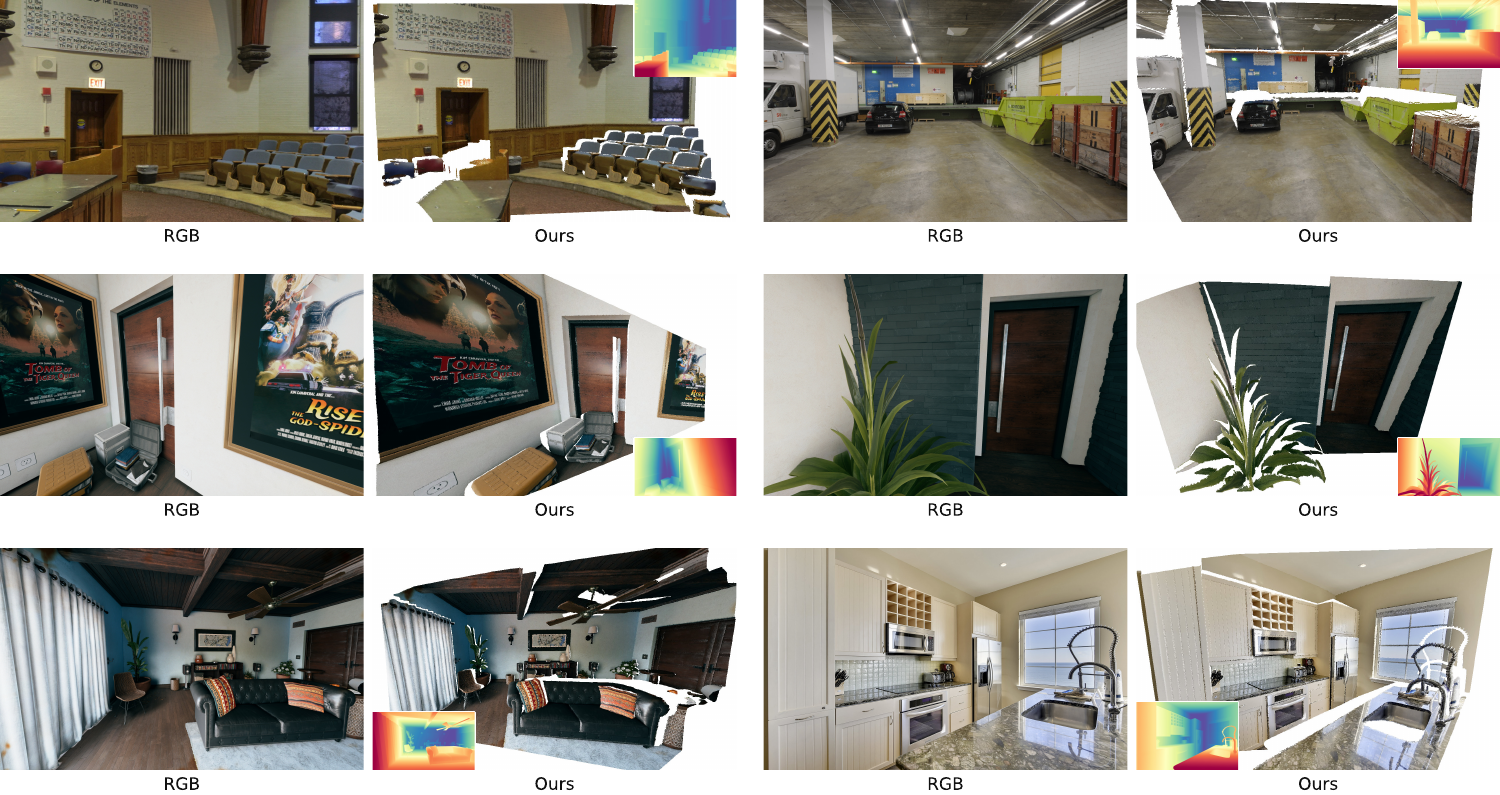}
    \caption{\textbf{Point cloud results of our method.}}
    \label{fig:supp_point_cloud}
\end{figure}

\begin{figure}[t]
    \centering
    \includegraphics[width=\linewidth]{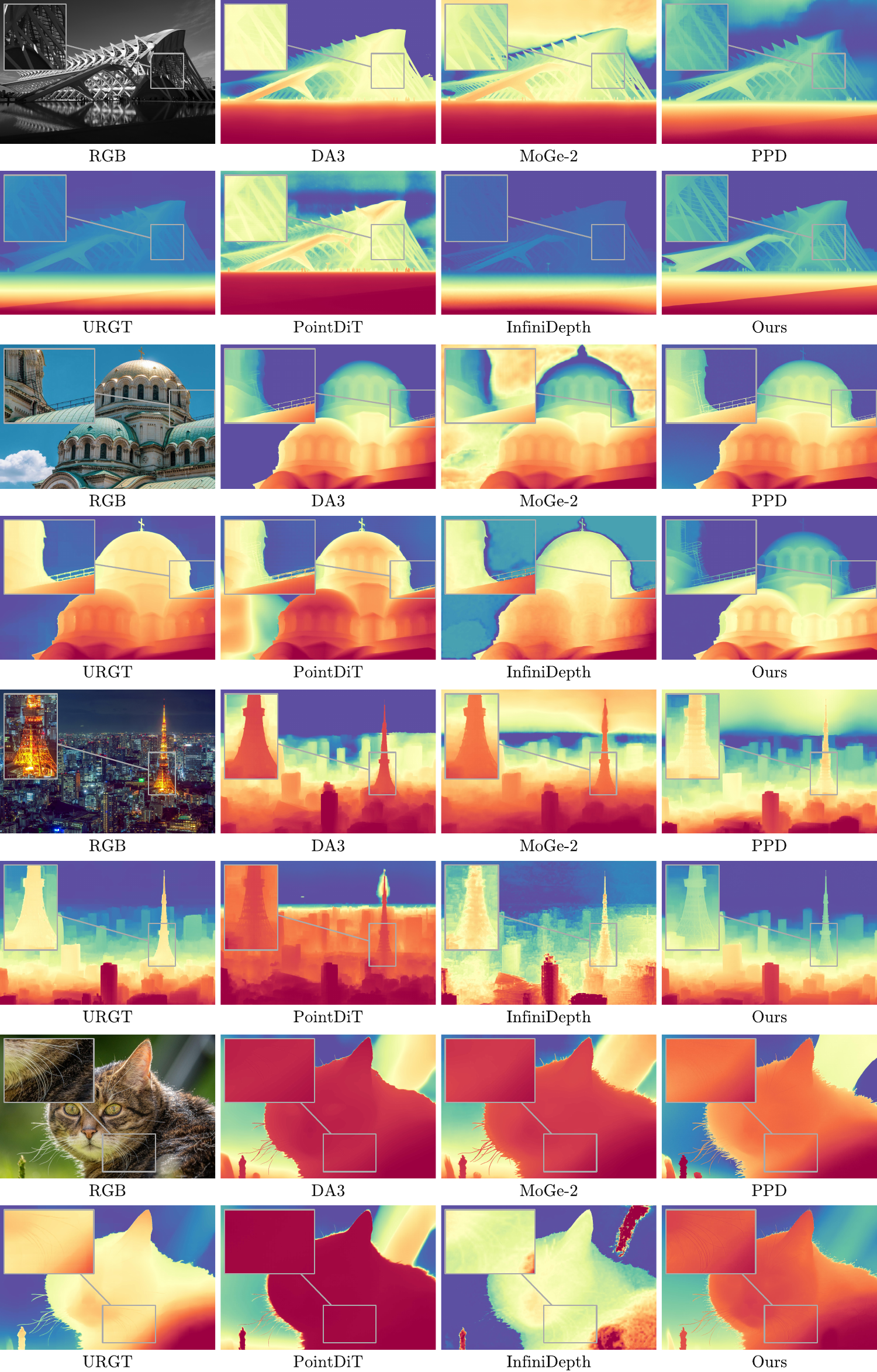}
    \caption{\textbf{More visual results (1).}  Zoom in for better visibility.}
    \label{fig:supp_visual_1}
\end{figure}

\begin{figure}[t]
    \centering
    \includegraphics[width=\linewidth]{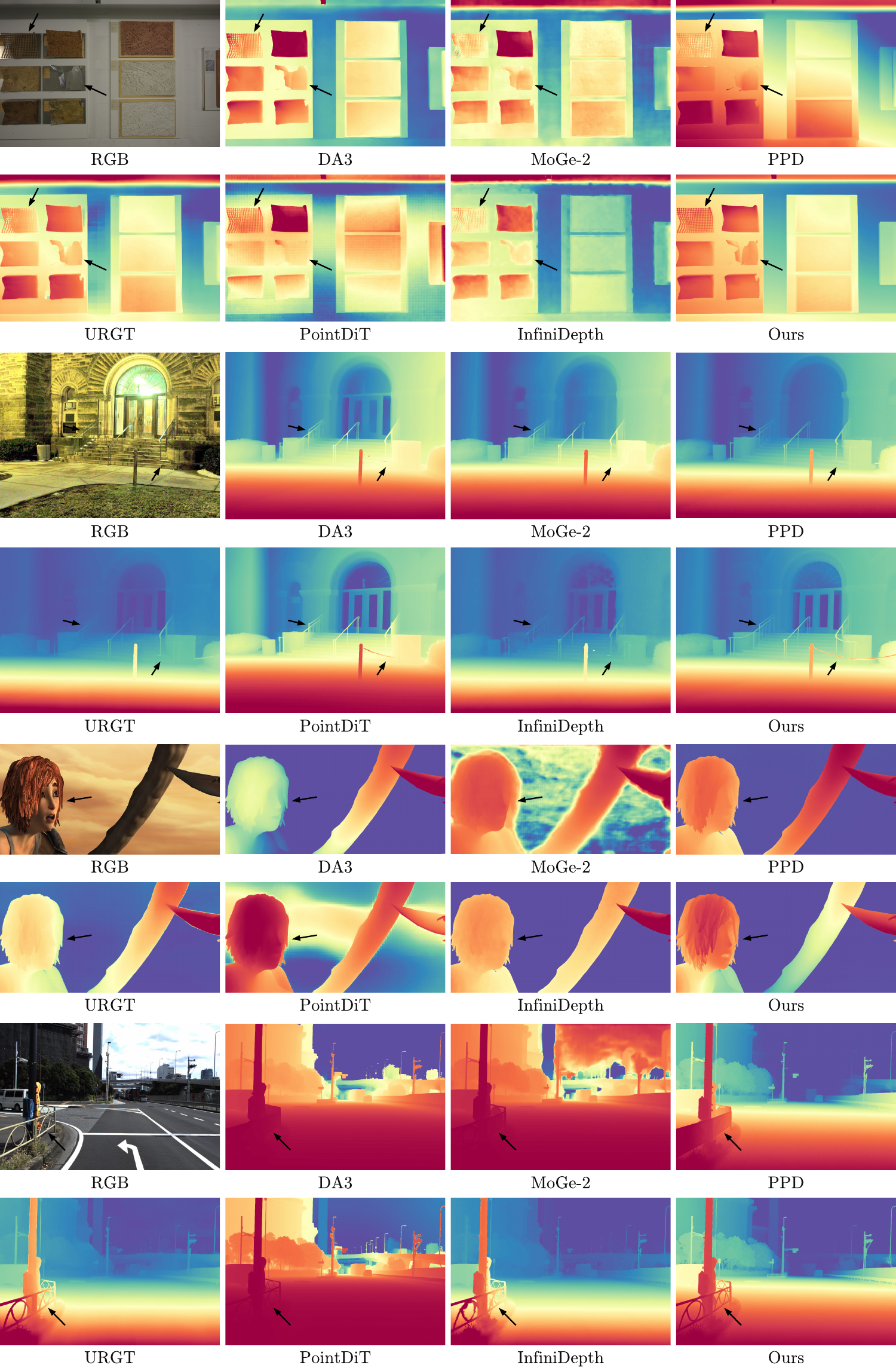}
    \caption{\textbf{More visual results (2).} Zoom in for better visibility.}
    \label{fig:supp_visual_2}
\end{figure}

\begin{figure}[t]
    \centering
    \includegraphics[width=\linewidth]{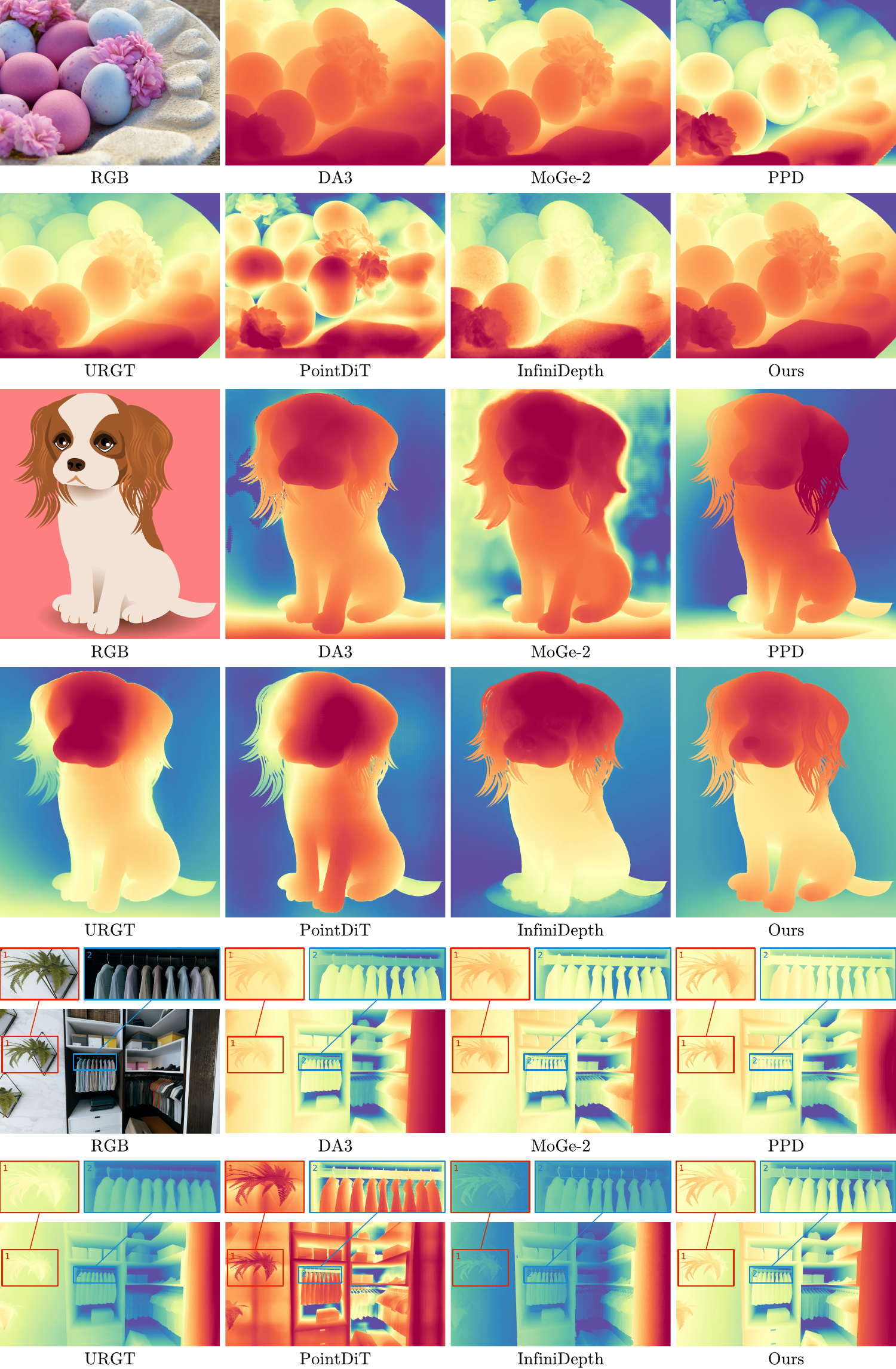}
    \caption{\textbf{More visual results (3).} Zoom in for better visibility.}
    \label{fig:supp_visual_3}
\end{figure}

Figure~\ref{fig:supp_point_cloud} presents point-cloud reconstructions from our predicted depth maps, obtained using the same depth-to-point-cloud conversion procedure as PPD~\citep{xu2026pixel}.
Figures~\ref{fig:supp_visual_1}, \ref{fig:supp_visual_2}, and~\ref{fig:supp_visual_3} provide additional qualitative comparisons with other methods, further illustrating our method's ability to recover fine geometric details while preserving overall scene structure.

\begin{table*}[t]
  \centering
  \caption{Component-wise runtime (ms$\downarrow$) and peak GPU memory
    (GB$\downarrow$) analysis. All runtime results are evaluated on an NVIDIA
    RTX 4090 GPU. CPM denotes the continuity-promoting module.}
  \label{tab:efficiency-components}
  \small
  \begin{tabular*}{\textwidth}{@{\extracolsep{\fill}}l*{4}{rr}@{}}
    \toprule
    \multirow{2}{*}{Component}
      & \multicolumn{2}{c}{$512\times512$}
      & \multicolumn{2}{c}{$1024\times1024$}
      & \multicolumn{2}{c}{$2048\times2048$}
      & \multicolumn{2}{c}{$2160\times3840$} \\
    \cmidrule(lr){2-3}\cmidrule(lr){4-5}
    \cmidrule(lr){6-7}\cmidrule(l){8-9}
      & Time & Memory
      & Time & Memory
      & Time & Memory
      & Time & Memory \\
    \midrule
    VAE Encoder
      & 16 & 11.9 & 83 & 13.2
      & 424 & 18.0 & 1058 & 24.4 \\
    DiT
      & 102 & 11.8 & 349 & 12.4
      & 1733 & 14.7 & 4423 & 17.6 \\
    VAE Decoder
      & 34 & 12.1 & 159 & 13.8
      & 732 & 20.7 & 1668 & 29.8 \\
    \midrule
    Pixel Diffusion
      & 37 & 11.8 & 78 & 12.4
      & 487 & 14.8 & 1298 & 18.0 \\
    CPM
      & 2 & 11.7 & 7 & 11.9
      & 39 & 13.1 & 77 & 14.7 \\
    \bottomrule
  \end{tabular*}
\end{table*}

\subsection{Component-wise efficiency.}
Table~\ref{tab:efficiency-components} provides a component-wise analysis of the VAE encoder, FLUX.2 DiT, VAE decoder, PiD, and continuity-promoting module (CPM).
Beyond its advantages in depth quality, the pixel diffusion decoder reduces decoding cost compared with the VAE decoder, including at high output resolutions.
The continuity-promoting module further suppresses patch-boundary artifacts with modest computational overhead.

\end{document}